\documentclass[a4paper,UKenglish,cleveref, autoref, thm-restate]{tgdk-v2021}

\usepackage{blindtext}
\usepackage{booktabs}
\usepackage[table,dvipsnames]{xcolor}

\usepackage{amsmath,amssymb,amsthm}
\usepackage{todonotes}
\usepackage{mathtools}
\usepackage{adjustbox}
\usepackage{rotating}
\usepackage{paralist}
\usepackage{tabularx}

\usepackage{listings} 
\theoremstyle{definition}
\usetikzlibrary{patterns}
\usepackage[T1]{fontenc}
\usepackage{subcaption}
\usepackage{graphicx}
\usepackage{algorithm, algpseudocode}

\usepackage{tikz}
\usetikzlibrary{positioning, shapes, arrows.meta}
\usepackage{multirow}
\usepackage{xspace}
\usepackage{array}
\usepackage{url}
\usepackage{lipsum}
\usepackage[utf8]{inputenc}		
\usepackage{amssymb}
\usepackage{pifont}

\DeclareMathOperator*{\argmax}{argmax}

\newcommand{\triple}[3]{\ensuremath{(#1, #2, #3)}}

\newcommand{\entities}{\ensuremath{\mathcal{E}}\xspace}
\newcommand{\relations}{\ensuremath{\mathcal{R}}\xspace}

\newcommand{\kg}{\ensuremath{\mathcal{G}}\xspace}
\newcommand{\scoreFunc}{\phi}

\title{Resilience in Knowledge Graph Embeddings} 

\author{Arnab Sharma}{Department of Computer Science, Paderborn University, Germany }{arnab.sharma@uni-paderborn.de}{https://orcid.org/0000-0002-1825-0097}{}
\author{N'Dah Jean Kouagou}{Department of Computer Science, Paderborn University, Germany}{
ndah.jean.kouagou@upb.de}{https://orcid.org/0000-0002-4217-897X}{}
\author{Axel-Cyrille Ngonga Ngomo}{Department of Computer Science, Paderborn University, Germany}{axel.ngonga@upb.de}{https://orcid.org/0000-0001-7112-3516}{}

\authorrunning{J. Open Access and J.\,R. Public} 

\Copyright{Jane Open Access and Joan R. Public} 

\keywords{Knowledge graphs, Resilience, Robustness} 

\category{} 

\relatedversion{} 

\nolinenumbers 

\DateSubmission{XX.XX.XXXX}
\DateAcceptance{XX.XX.XXXX}
\DatePublished{XX.XX.XXXX}

\begin{document}

\maketitle

\begin{abstract}
In recent years, knowledge graphs have gained interest and witnessed widespread applications in various domains, such as information retrieval, question-answering, recommendation systems, amongst others. Large-scale knowledge graphs to this end have demonstrated their utility in effectively representing structured knowledge.  To further facilitate the application of machine learning techniques,  knowledge graph embedding (KGE) models have been developed.  Such models can transform entities and relationships within knowledge graphs into vectors. However, these embedding models often face challenges related to noise, missing information, distribution shift, adversarial attacks, etc. This can lead to sub-optimal embeddings and incorrect inferences, thereby negatively impacting downstream applications. While the existing literature has focused so far on adversarial attacks on KGE models, the challenges related to the other critical aspects remain unexplored. In this paper, we, first of all, give a unified definition of {\em resilience}, encompassing several factors such as generalisation, performance consistency, distribution adaption, and robustness. After formalizing these concepts for machine learning in general, we define them in the context of knowledge graphs. To find the gap in the existing works on resilience in the context of knowledge graphs, we perform a systematic survey, taking into account all these aspects mentioned previously. Our survey results show that most of the existing works focus on a specific aspect of resilience, namely robustness. After categorizing such works based on their respective aspects of resilience, we discuss the challenges and future research directions. 
\end{abstract}

\newcommand{\an}[1]{\todo[inline,color=yellow]{#1}}

\section{Introduction}
\label{sec:introduction}

In recent years, there has been significant progress in the construction and application of knowledge graphs (KGs). Many KGs, including Freebase~\cite{bollacker2008freebase}, DBpedia~\cite{AuerBKLCI07}, YAGO~\cite{suchanek2007yago}, and NELL~\cite{CarlsonBKSHM10}, have been developed and successfully implemented in various real-world applications. Due to their effectiveness in knowledge representation, 
KGs now find applications in various domains such as information retrieval \cite{DaltonDA14}, question answering \cite{FerrucciBCFGKLMNPSW10}, and recommendation systems \cite{WangZZLXG19}, amongst others. Typically, a KG serves as a structured depiction of knowledge, organized as a multi-relational graph where nodes stand for entities or concepts, and edges signify relationships between them~\cite{hogan2021knowledge}. 
Knowledge therein is represented using assertions--that are model statements  (which could in some cases be real-world facts)--in the form of triples denoted as 
\( (h, r, t) \), where \( h \) and \( t \) correspond to the head and tail entities respectively, and \( r \) represents the relationship between them. For instance, the fact `Biden is the president of USA' can be represented in a KG as \( ( \texttt{Biden}, \texttt{presidentOf}, \texttt{USA} ) \).

KGE involves transforming the entities and relations within a KG into vectors~\cite{bordes2013translating,TYL19,dettmers2018convolutional,wang2021survey,zamini2022review}. This transformation makes computational operations more feasible, allowing machine learning and deep learning techniques to be applied to extract insights from the KG. Consequently, an effective KGE model should aim to preserve the properties and semantics inherent in the original KG. Based on the type of KGE models, entities and relations are commonly embedded  in $d$-dimensional vector spaces $\mathbb{V}$ such as $\mathbb{R}^d$ (real numbers)~\cite{bordes2013translating}, $\mathbb{C}^d$ (complex numbers)~\cite{trouillon2016complex}, or even $\mathbb{H}^d$ (quaternions)~\cite{chami2020low}.
 
Despite their effectiveness in capturing complex relationships between entities of KGs and facilitating various downstream tasks, KGE models can be vulnerable to various attacks. Since these models rely heavily on the observed connections in a given graph, noise or missing information can lead to sub-optimal embeddings and potentially incorrect inferences. For instance, the presence of incorrect triples (in the sense of non-conformity with an ontology, or wrong assertions) might lead to poorly performing KGE models on certain downstream tasks.
KGE models might also struggle to generalize to out-of-distribution or unseen data, e.g., when the underlying data distribution changes or when encountering new entities or relations. Since KGs often contain sensitive and critical information pertaining to individuals or organizations, this might give rise to potential security vulnerabilities. For instance, an attacker might subtly alter the relation between entities or introduce fictitious entities and relationships that distort the model's understanding of the graph and make the KGE model learn {\em poisoned} embeddings. Such adversarial attacks on KGE models can take various forms, such as adding, deleting, or modifying triples within the knowledge graph where such perturbations are often minimal and crafted to exploit vulnerabilities in the embedding process. 
Due to the usage of KGE models in various downstream tasks, such adversarial attacks can cause potential disruptions in these tasks, for instance in 
\begin{compactenum}
    \item \textbf{question answering:} adversarial modifications can cause KGE models to produce incorrect or manipulated answers or fail to retrieve relevant information,
    \item \textbf{recommendation systems:} the embeddings can be poisoned to promote certain items unfairly, leading to biased or irrelevant recommendations,
    \item \textbf{information extraction:} adversarial perturbations can result in inaccurate extractions of facts, affecting downstream applications like content summarization or data integration, amongst others.
\end{compactenum}
Therefore, to reliably use KGE models in downstream tasks, there is a need to develop models that can function without any potential disruption of their performance even in the presence of such adversarial conditions.



Although the aforementioned challenges pose potential threats to the use of KGE models in critical downstream tasks, current efforts to deal with these challenges still remain infancy. The existing literature mostly contains works addressing challenges related to noisy data, distribution shifts, and adversarial attacks in the context of graph neural networks~\cite{XL0C22,inv-shift,Zhang0ZLQ022,FanWSCW24}. 
So far, works considering KGE models mostly focused on performing adversarial attacks on them~\cite{ZhangZGMSL019, BhardwajKCO20, BhardwajKCO21, YouSDZPYF23}. 
The core idea behind these attacks is to target specific facts and modify the KGE model to either increase or decrease their plausibility scores. These scores reflect the likelihood of a fact being true: higher scores imply higher probability, while lower scores imply lower probability. For instance,  if (\texttt{Biden}, \texttt{PresidentOf}, \texttt{USA})  is selected as the target triple, one type of adversarial attack would be to make the underlying KGE model assign a low plausibility score to it. In this case, such attacks are typically dealt with via a min-max optimization function, where the objective is to minimize the inclusion/deletion of adversarial/existing triples in/from the underlying KG~\cite{YouSDZPYF23}. Simultaneously, the attacker aims to maximize the objective function, which involves either increasing or decreasing the plausibility of a targeted fact being true. 


Since KGs are used in many safety-critical environments, safeguarding sensitive information and preserving user privacy are paramount considerations in deploying KGE models in real-world settings. Furthermore, we need to enable KGE models to adapt to dynamic environments and evolving data distributions to enhance their resilience to concept drift and temporal changes. Therefore, in this work, we first of all propose the concept of {\em resilience} in the context of ML, and further extend the definition for KGE models. 
We aim to bridge the gap in resilience literature on KGE from a holistic perspective that considers the diverse facets of robustness, adaptability, distribution shift, and consistency, amongst others. By addressing these aspects comprehensively, researchers can propel the development of resilient KGE models that not only excel in performance metrics but also demonstrate stability and reliability in real-world applications. 
Precisely, we give a generic formal definition of resilience in ML models considering 
\begin{inparaenum}[(i)]
\item generalization consistency, 
\item domain adaptation, 
\item performance consistency, 
\item robustness, and
\item missing entry handling. 
\end{inparaenum}
We then discuss these aspects of resilience in the context of KGE models. To this end, we survey the works on KGE models considering the aforementioned aspects of resilience. Specifically,  we provide a survey of works studying the resilience of KGE models in any of the aspects from (i)--(v). After discussing these works, we highlight possible challenges and suggest future work directions.

This paper is organized as follows. Section~\ref{sec:foundation} formalizes the notions of KGs and KGE models. The definition of resilience and its aspects are discussed in Section~\ref{sec:resiliency-definition}. Section~\ref{sec:paper-collection} describes the methodology regarding the collection of papers. Existing works discussing aspects of resilience are presented in Section~\ref{sec:reslience-KGE}. Section~\ref{sec:robustness} presents and discusses different aspects of robustness. Section~\ref{sec:challenges} highlights existing challenges and potential future work directions, and Section~\ref{sec:conclusion} concludes the paper.

\section{Foundations}\label{sec:foundation}

A knowledge graph is a collection of assertions that describe a domain of interest. In this paper, we consider knowledge graphs composed of {\em triples} $\triple{h}{r}{t} \in  \entities \times \relations \times \entities$, where $\entities$ is a discrete set of entities and $\relations$ is a discrete set of relations. Therefore, KGs are representations of information in a discrete space. More formally, a KG is defined as a set of triples $\kg:= \{ \triple{h}{r}{t}\in \entities \times \relations \times \entities\}$, where \entities\ and \relations\ stand for a set of entities and a set of relations~\cite{dettmers2018convolutional,balavzevic2019tucker}.
However, such a graph cannot be processed easily. Hence, KGE algorithms have been developed to represent the knowledge of a KG in a continuous, low dimensional embedding space. Essentially, the KGE models learn continuous vector representations tailored towards link prediction which can be defined as the parameterized scoring function. Let $\mathbb{V}$ denote a normed-division algebra, e.g. $\mathbb{R},\mathbb{C},\mathbb{H}$, or $\mathbb{O}$~\cite{balavzevic2019hypernetwork,demir2021hyperconvolutional,yang2014embedding,trouillon2016complex,zhang2019quaternion}.
A KGE model of a KG comprises 
entity embeddings $\mathbf{E} \in \mathbb{V}^{|\mathcal{E}| \times d_e}$ and relation embeddings $\mathbf{R} \in \mathbb{V}^{|\mathcal{R}| \times d_r}$, where $d_e$ and $d_r$ are the size of the embedding vectors. In the following, we use $d$ as size for all embedding vectors, as it has been shown that $d_e = d_r$ holds for many types of models~\cite{nickel2015review}. Throughout this paper, we will denote  embedding vectors with bold fonts, for instance, the embedding of $h$, $r$, and $t$ will be denoted as $\mathbf{h}$, $\mathbf{r}$, and $\mathbf{t}$, respectively.

Given a KG $\mathcal{G}\subseteq \entities \times \relations \times \entities$, the goal of a KGE model is to learn continuous vector representations for entities and relation types in $\mathcal{G}$ such that these representations can be used to recover all the facts in $\mathcal{G}$.
Most KGE approaches are tailored towards link prediction~\cite{chami2020low,hogan2021knowledge}, i.e., their scoring function is $\scoreFunc_\Theta: \entities \times \relations \times \entities \rightarrow \mathbb{R}$,
where $\Theta$ denotes parameters and often comprises $\mathbf{E}$, $\mathbf{R}$, and additional parameters (e.g., affine transformations, batch normalizations, convolutions).
Given an assertion in the form of a triple $\triple{h}{r}{t} \in \entities \times \relations \times \entities$, a prediction $\hat{y}:=\scoreFunc_\Theta\triple{h}{r}{t}$ signals the likelihood of $\triple{h}{r}{t}$ being true~\cite{dettmers2018convolutional,wang2021survey,zamini2022review}. Therefore, KGE models are learned in such a way that the scoring function assigns a higher score to the triples that exist in the KG compared to the non-existing ones.

Since KGs contain triples which represent the existing facts only, to learn a KGE model effectively, non-existing facts, i.e., {\em negative} facts often need to be incorporated into the learning process. For that, a technique called negative sampling is used to generate a number of false facts or negative triples. To this end, Bordes et al. \cite{bordes2013translating} proposed a negative sampling technique by perturbing an entity in a randomly sampled triple from the KG.
In this setting, a triple $\triple{h}{r}{t} \in \kg$ is considered as a positive example, whilst 
$\{ \triple{h}{r}{x} |x \in \entities \wedge \triple{h}{r}{x} \not \in \kg\} \cup \{ \triple{x}{r}{t} |x \in \entities \wedge \triple{x}{r}{t} \not \in \kg \}$ is regarded as the set of possible candidate negative triples corresponding to $\triple{h}{r}{t}$. During training, $k$ negative triples are constructed for every  correct triple. Consequently, a mini-batch $\mathcal{B}$ consists of $m$ positive triples, $k \times m$ negative triples, and the respective $m + k \times m$ binary labels.

\section{Resilience}\label{sec:resiliency-definition}
As mentioned beforehand, resilience is a term that is frequently used when engineering systems, more specifically in the context of building fault-tolerant systems \cite{Strigini12}. 
In those systems, \emph{resilience refers to the ability of a system to maintain its functionality and performance in the face of faults, failures, disruptions, or adverse conditions}. In other words, a resilient system is capable of detecting, mitigating, and recovering from faults or failures, ensuring continuous operation and minimal impact on its overall performance and availability. Therefore, in~\cite{BergerERDHH22}, the authors defined the resilience of a system using 
\begin{compactenum}
\item \textbf{availability}, i.e., the readiness for correct service, 
\item \textbf{reliability}, i.e.,
the probability of performing correctly for a period of time, 
\item \textbf{safety}, i.e., the robustness against adversarial manipulations, 
\item \textbf{integrity}, i.e., the absence of improper system altercation, and 
\item \textbf{maintainability}, i.e., the ability to undergo modifications and repairs. 
\end{compactenum}
While the typical definition of resilience in fault-tolerant systems provides a useful starting point for understanding resilience in the machine learning domain, it needs to be extended and adapted to account for the unique characteristics, challenges, and considerations inherent in machine learning models and systems. More specifically, for ML models, resilience cannot be defined by using these parameters directly since they do not capture the typical data-driven workflow that is used in ML. For this, we need to consider other factors such as {\em consistent} behavior, {\em distribution shift}, robustness, amongst others 

To define resilience formally, we start with some basic formalization.
Let us consider an ML model as a function $f$ which takes as input $x$ coming from a specific distribution $\mathcal{D}$. We define two types of distributions from where the data might come, the source distribution which is defined as $\mathcal{D}_s$ from where the training data comes, and the target distribution $\mathcal{D}_t$ on which the model would typically operate. The sets of values corresponding to the distributions $\mathcal{D}_s$ and $\mathcal{D}_t$ can be defined as $\mathcal{X}_s$ and $\mathcal{X}_t$, respectively. $\mathcal{H}(\mathcal{D}_s, \mathcal{D}_t)$ defines a divergence measure between the two distributions $\mathcal{D}_s$ and $\mathcal{D}_t$. 
Furthermore, we define $\mathcal{L}_f$ as the {\em loss function} of the model measuring the model's performance on a set of data instances. Note that, the loss function can be of any type, however, our definition is independent of it. An ML model $f$ is said to be resilient if it conforms to the following five constraints:

\begin{description}
    \item[Generalization consistency] This condition basically corresponds to the ability of the model to generalize consistently across different distributions of data. This can be formally defined as,
    \begin{equation}
          \forall\ \mathcal{D}_s, \mathcal{D}_t,  \: |\mathbb{E}_{\mathcal{D}_s} (\mathcal{L}_f) - \mathbb{E}_{\mathcal{D}_t}(\mathcal{L}_f)| \leq \epsilon
    \end{equation}
    where $\epsilon$ defines a threshold that basically bounds the difference between the average losses on the training data distribution $\mathcal{D}_s$ and the target data distribution $\mathcal{D}_t$. Xu et al.~\cite{XuM10} defined this as the robustness property of the learning algorithms where they argued that a robust algorithm should achieve similar performance on the training and testing data that are {\em close} in some sense; which basically corresponds to the robust optimization problem. However, in connection to resilience, we define this as the {\em consistency} property over the generalization of the model $f$. To this end, we simply say that the loss occurring on the data instances taken from the target distribution might differ only by a threshold $\epsilon$ from the loss occurring on the instances of the source distribution. Furthermore, in this definition, we are not concerned with whether a model $f$ achieves high accuracy or low loss on the training data; with generalization consistency, we aim to signify that the model's performance should not vary drastically between the training and test data distributions.
 
    In the context of knowledge graph embedding models, generalization consistency refers to the model's ability to meaningfully construct embeddings for unseen entities or relations, and accurately predict missing links between entities based on the learned patterns from the training data. 
    Assuming $\mathcal{L}_{\phi_{\Theta}}$ is a loss function which can be used to train the parameterized embedding model $\phi_{\Theta}$, generalization consistency can be defined as
    \begin{equation}
     \forall\ \mathcal{D}_\mathcal{G}, \mathcal{D}_{\mathcal{G'}},  \: |\mathbb{E}_{{\mathcal{G}}} (\mathcal{L}_{\phi_{\Theta}}) - \mathbb{E}_{{\mathcal{G'}}}(\mathcal{L}_{\phi_{\Theta}})| \leq \epsilon 
    \end{equation} 

    where $\mathcal{D}_{\mathcal{G}}$ and $\mathcal{D}_{\mathcal{G'}}$ refer to the distribution of the training knowledge graph's data and that of the test knowledge graph's data, respectively.

    \item[Distribution adaption] This corresponds to the model's ability to adapt to a target domain (i.e., test data distribution) without significantly compromising its performance as achieved on the source domain (i.e., training data distribution). This can be defined as follows
    \begin{equation}
     \forall\ \mathcal{D}_s, \mathcal{D}_t:\  \mathcal{H}(\mathcal{D}_s, \mathcal{D}_t) \leq \delta \Rightarrow  
    | \mathbb{E}(\mathcal{L}_f)  - \mathbb{E}(\mathcal{L}_f) | \leq \epsilon 
    \end{equation}
    where $\mathcal{H}(\mathcal{D}_s, \mathcal{D}_t)$ defines any divergence measure such as maximum mean discrepancy (MMD)~\cite{SmolaGSS07}, Kullback-Leibler (KL) divergence~\cite{kullback1951information}, or Wasserstein distance~\cite{wasserstein1969markov}. Informally, if the distributions $\mathcal{D}_s$ and $\mathcal{D}_t$ are different with a bound $\delta$, then the average prediction losses on the data instances in these distributions must not differ more than $\epsilon$. Note that, the distributional mismatch between the training and test data has been studied in many settings, for instance, in~\cite{dataset-shift, shimodaira2000improving, HuangSGBS06, BickelBS07, DworkHPRZ12, abs-2005-12914} and as pointed out by the authors in~\cite{AgarwalZ22} most of these works assume the {\em covariate shift}
    where only the distribution of class labels differs between the training and test distributions. There exist some works such as~\cite{Ben-DavidBCKPV10, FeigeMS15, Csurka20} which consider shift of generic data distributions, however, none of them consider this as part of the resilience of ML models.
    For KGE models, distribution adaptation refers to a model's ability to adjust its parameters to account for changes in a given knowledge graph. When new entities, relation types, or new links are added to (or removed from) a given knowledge graph, the resulting graph data distribution might deviate from the initial one.
    In this case, the KGE model's adaptation to this distribution change can be formally defined as follows
    \begin{equation}
         \forall\ \mathcal{D}_{\mathcal{G}}, \mathcal{D}_{\mathcal{G'}}:\ \mathcal{H}(\mathcal{D}_{\mathcal{G}}, \mathcal{D}_{\mathcal{G'}})\leq \delta \Rightarrow |\mathbb{E}_{{\mathcal{G}}} (\mathcal{L}_{\phi_{\Theta}}) - \mathbb{E}_{{\mathcal{G'}}}(\mathcal{L}_{\phi_{\Theta}})| \leq \epsilon
         \label{eq:hd}
    \end{equation}
    Note that, in the context of graphs, a distribution shift refers to a change in the statistical distribution of the graph data. This can manifest in different ways, such as \begin{compactenum}
        \item \textbf{node feature distribution shift:} This occurs when the distribution of node attributes or features changes over time or across different subsets of the graph. For example, in a knowledge graph representing entities and their attributes (e.g., people and their professions), a node attribute shift could involve changes in the distribution of professions among individuals over time or across different subsets of the graph. Nodes may furthermore be added to or removed from the knowledge graph, leading to changes in the overall node distribution. This could happen, for instance, when new entities are discovered or when outdated entities are removed from the knowledge graph.

        \item \textbf{node degree shift:} This happens when some relationships between entities are removed (e.g., two entities that were previously friends are no longer friends) or added, e.g., (an entity gets married to another entity). It could also be the case that new entities are introduced but with little to zero links to other entities in the graph. When such changes in relationships between entities are significant, the average degree of nodes in the considered knowledge graph might also shift.

        \item \textbf{edge feature distribution shift:} This refers to changes in the properties or attributes associated with the relationships (edges) between nodes in the knowledge graph. For example, in a knowledge graph representing relationships between entities (e.g., co-authorship relationships between researchers), an edge attribute shift could involve changes in the publication venues or collaboration patterns over time.  New relationships may further be established or existing relationships may be removed from the knowledge graph, leading to changes in the edge distribution. This could occur due to the emergence of new relationships or the obsolescence of existing ones.

        \item \textbf{graph structure shift:}  This involves alterations in the overall structure or topology of the knowledge graph, including changes in connectivity patterns between nodes, changes in node/edge attributes (e.g., many entities and relationships in the reference knowledge graph now have textual descriptions), and changes in entity type hierarchies. 
        For example, in a knowledge graph representing hierarchical relationships (e.g., taxonomy or ontology), changes in the hierarchy or the addition of new branches can lead to structural shifts. Changes to the schema or ontology of the knowledge graph, such as the addition, modification, or removal of entity types, relationship types, or property types, can also constitute graph structure shifts. These changes may reflect updates in domain knowledge or evolving data modeling requirements.
    \end{compactenum}

    
    \item[Performance consistency] This corresponds to the model's ability to perform consistently across different instances or subsets of a data distribution. Typically, this distribution could be the target distribution $\mathcal{D}_t$ where the data instances come from in the model deployment phase. Formally, consistency can be defined as follows:

    \begin{equation}
        \forall\ \mathcal{S}, \mathcal{S'} \subseteq \mathcal{D}_t\ :
    |\mathbb{E}_{\mathcal{S}}(\mathcal{L}_f) - \mathbb{E}_{\mathcal{S}'}(\mathcal{L}_f) | \leq \epsilon.
    \end{equation}
    Here we enforce that, for any two non-empty subsets $\mathcal{S}$ and $\mathcal{S'}$ from the distribution $\mathcal{D}_t$, the expected losses achieved on the two sets differ at most only by some parameter $\epsilon$~\footnote{Note that, here we have considered a strong notion of consistency, however, a weaker notion can also be chosen where the subsets must follow some specific rules.}. 
    If the observed differences between the losses  are statistically significant (e.g., greater than $\epsilon$), it indicates that the model's performance varies consistently across different subsets of the data, suggesting potential limitations or biases in the model. On the other hand, if the observed differences are not statistically significant, it suggests that the model's performance remains consistent across subsets, providing greater confidence in its resilience. Furthermore, this measure of consistency is different from the generalization consistency in the sense that herein we consider uniform performance across different sub-spaces of the same distribution space, whereas in case of generalization consistency, two different distributions are considered.

    Performance consistency for the KGE models refers to the model's ability to maintain consistent performance across different instances or subsets of the knowledge graph data distribution, particularly when deployed in real-world applications where the distribution of incoming triples may vary. In other words, the KGE model should demonstrate resilience to variations in the distribution of knowledge graph data encountered during deployment, ensuring that its performance remains reliable and predictable across different scenarios. This consistency is crucial for maintaining the effectiveness and reliability of the model in real-world applications where the knowledge graph would evolve over time or across different contexts.

    \item[Robustness] This aspect of resilience focuses on the model's stability with respect to some small changes in the input. In the literature, two versions of robustness are generally considered, namely {\em local} and {\em global} robustness~\cite{GoodfellowSS14, SeshiaDDFGKSVY18}. Informally, local robustness corresponds to a single point $x$, and requires any points within a specific distance of $\Delta$  to $x$ to be classified as the same as the former. More formally, this can be expressed as
    \begin{equation}
        \forall\ x, x',\ ||x-x'||_p \leq \Delta \Rightarrow ||f(x) - f(x')||_p \leq \epsilon. 
    \end{equation}
    On the other hand, Seshia et al.~\cite{SeshiaDDFGKSVY18} defined global robustness considering all the points within a specific distribution $\mathcal{D}$. In other words, for every point $x$ within a considered distribution, any other point $x'$ which is within $\Delta$ distance from $x$ should be classified as the same class as $x$. This can be formally defined as 
    \begin{equation}
        \forall\ x, x' \in \mathcal{D},\ ||x-x'||_p \leq \Delta \Rightarrow ||f(x) - f(x')|| \leq \epsilon
    \end{equation}
    Note that, in the literature, robustness is more often associated with the idea of local robustness for a single point or a set of points. Thus, in defining resilience, we would primarily consider local robustness property of ML models.  

    In the context of KGE models, we can adapt the concept of local robustness to refer to the model's ability to produce consistent embeddings for entities or relations that are similar in the graph structure. Informally, local robustness in this context would correspond to: any entity (respectively, relation) within a specific neighborhood of an entity $h$ (respectively a relation $r$), defined by a distance metric, should have an embedding that is similar to that of $h$ (respectively $r$). More formally, for an entity or 
    a relation $x$ in the knowledge graph, and for any other entity or relation $x'$ within a specific distance $\Delta$ of $x$ in the graph structure, the embeddings produced by the KGE model, say $\mathbf{x}$ and $\mathbf{x}'$) should be similar, with their distance in the embedding space bounded by \(\epsilon\). Given a knowledge graph $\mathcal{G}$, this idea of robustness can be formally defined as
    \begin{equation}\label{equation:robustness}
        \forall\ x, x'\in \mathcal{G}, d_\mathcal{G}(x, x') \leq \Delta \Rightarrow d_{Emb}(\mathbf{x}, \mathbf{x'}) \leq \epsilon,
    \end{equation}
where $d_\mathcal{G}: \mathcal{G}\times \mathcal{G}\rightarrow \mathbb{R}_+$ is a distance on the graph $\mathcal{G}$, e.g., Adamic-Adar index, Katz similarity, or Common Neighbors, and $d_{Emb}$ a distance function in the embedding space, e.g., the Euclidean distance. $\epsilon$ is a threshold that limits the allowable difference between embeddings to ensure local robustness. 

Considering the structure of the KG, and its effect on the learned embeddings, robustness might also correspond to the ability of the KGE models to be invariant against a certain level of noise introduced in the KG, compared to a {\em clean} KG. More specifically, the performance of a robust KGE model should not degrade considerably when noise is prevalent in the KG. Consider $\mathcal{G}$ as a clean KG and $\mathcal{G}'$ as a noisy KG, where the latter is obtained by adding $\delta$ amount of noise to the KG $\mathcal{G}$, i.e., $\mathcal{G}' = \mathcal{G} + \delta$.~\footnote{Here, $+$ is not the usual addition, but a perturbation operator instead.} Then the non-adversarial robustness can be defined as 
\begin{equation}
   \frac{ \mathbb{E}_{{\mathcal{G}}} (\mathcal{L}_{\phi_{\Theta}})}{\mathbb{E}_{{\mathcal{G'}}} (\mathcal{L}_{\phi_{\Theta}})} \approx 1, 
\end{equation}
where $\mathbb{E}_{\mathcal{G}}(\mathcal{L}_{\phi_{\Theta}})$, $\mathbb{E}_{\mathcal{G}'}(\mathcal{L}_{\phi_{\Theta}})$ denote the expected loss of the embedding model $\phi_{\Theta}$ on the graphs $\mathcal{G}$ and $\mathcal{G}$, respectively. This implies that the performance of the KGE models should remain almost the same even when $\delta$ amount of noise is present in $\mathcal{G}$. Note that we assume the expected loss not to be zero, as it is often the case in most machine learning tasks.

\item[Stability to incomplete input] This aspect of resilience deals with the model's ability to handle missing values, more specifically, maintain accurate predictions despite the presence of missing values in the input features. We can express this as follows:
        \begin{equation}
            \forall\ x, x^*, |x| > |x^*|\ \land\ \text{Sim}(x,x^*) \leq \delta \Rightarrow || f(x) - f(x') || \leq \epsilon.
        \end{equation}
     Herein $\text{Sim}(x,x^*)$ measures the similarity between two vectors $x$ and $x^*$ with unequal numbers of elements. One such similarity measure could be the cosine similarity, which is often used for comparing the similarity between vectors in high-dimensional spaces. The cosine similarity measures the cosine of the angle between two vectors and is defined as the dot product of the vectors divided by the product of their magnitudes. Elements that are missing in one vector but present in the other are effectively treated as zeros in the dot product calculation. Another approach is to use similarity measures that explicitly handle missing values, such as the Jaccard similarity or the Pearson correlation coefficient with the imputation of missing values. Thus, the similarity measure can be quite flexible and will potentially depend on the domain (for e.g., image classification, graph data, and others), and hence, we do not fix the Sim() function. Depending on this function and the domain of the application, the bound $\delta$ will also change, however, not drastically. We believe this requires further study, which is out of scope for this paper.
    
\end{description}

\section{Paper Collection Methodology}\label{sec:paper-collection}

To discuss resilience in KGs and KGE models, in this work, we further review the existing works done to this end. While doing a literature survey of such works, we adhere to specific inclusion criteria for compiling papers for our review. If a paper satisfies any one or more of the following criteria, it is considered for inclusion:
\begin{compactenum}
    \item the paper introduces or discusses the overarching concept of any related aspect outlined in Section~\ref{sec:resiliency-definition}.
    \item the paper proposes an approach, study, or tool/framework aimed at developing resilient or robust KGE models.
    \item the paper introduces a set of measurement criteria applicable for defining resilience of KGE models or KGs.
\end{compactenum}
\noindent
We briefly discuss some papers focusing solely on using KGs to make resilient systems, however, we do not delve into detail on this.  To comprehensively gather papers across various research domains, we initiated our search process by employing precise keyword queries on prominent scientific databases such as Google Scholar, DBLP, and arXiv. The keywords that we searched for are detailed in the \textbf{Keywords} column in Table~\ref{table:data}. We conducted searches across the three repositories until 23.09.2024, aiming to encompass a broad spectrum of literature. The specifics of the paper collection outcomes are outlined in Table~\ref{table:data}. It is observed that the papers obtained from Google Scholar and arXiv were subsets of those gathered from DBLP. Therefore, we solely present the results obtained from DBLP. Furthermore note that apart from the papers that discuss resilience in KGE models, or in KGs, we also report the results here where any of the aspects of resilience (as described in Section~\ref{sec:resiliency-definition}) are discussed in the body of the paper.



\begin{table}[t]
    \caption{Paper query results. Here, ``Body" represents the main content of a paper. Numbers correspond to the number of occurrences of a keyword.}
    \centering
    \small
    \label{table:data}
    \begin{tabular}{l@{\hspace{3mm}}c@{\hspace{3mm}}c@{\hspace{3mm}}c@{\hspace{3mm}}c@{\hspace{3mm}}c}
    \toprule
    \textbf{Keywords} & Title & Body \\
    \midrule
     resilience in knowledge graphs         & 4      & 0  \\
     resilience in knowledge graph embedding models     & 2      & 6 \\
     \midrule
     generalization consistency in/of knowledge graphs & 0 & 0 \\
     generalization consistency in/of knowledge graph embedding models & 0 & 0 \\
     domain adaption in/of knowledge graph embedding models & 0 & 2 \\
     distribution shift of knowledge graphs & 1 & 0 \\
     performance consistency of knowledge graph embedding models & 1 & 0 \\
     robustness of knowledge graph embedding models & 8 & 16\\

    \bottomrule
    \end{tabular}
\end{table}

\begin{figure*}[t]
    \centering
    \includegraphics[scale=0.80]{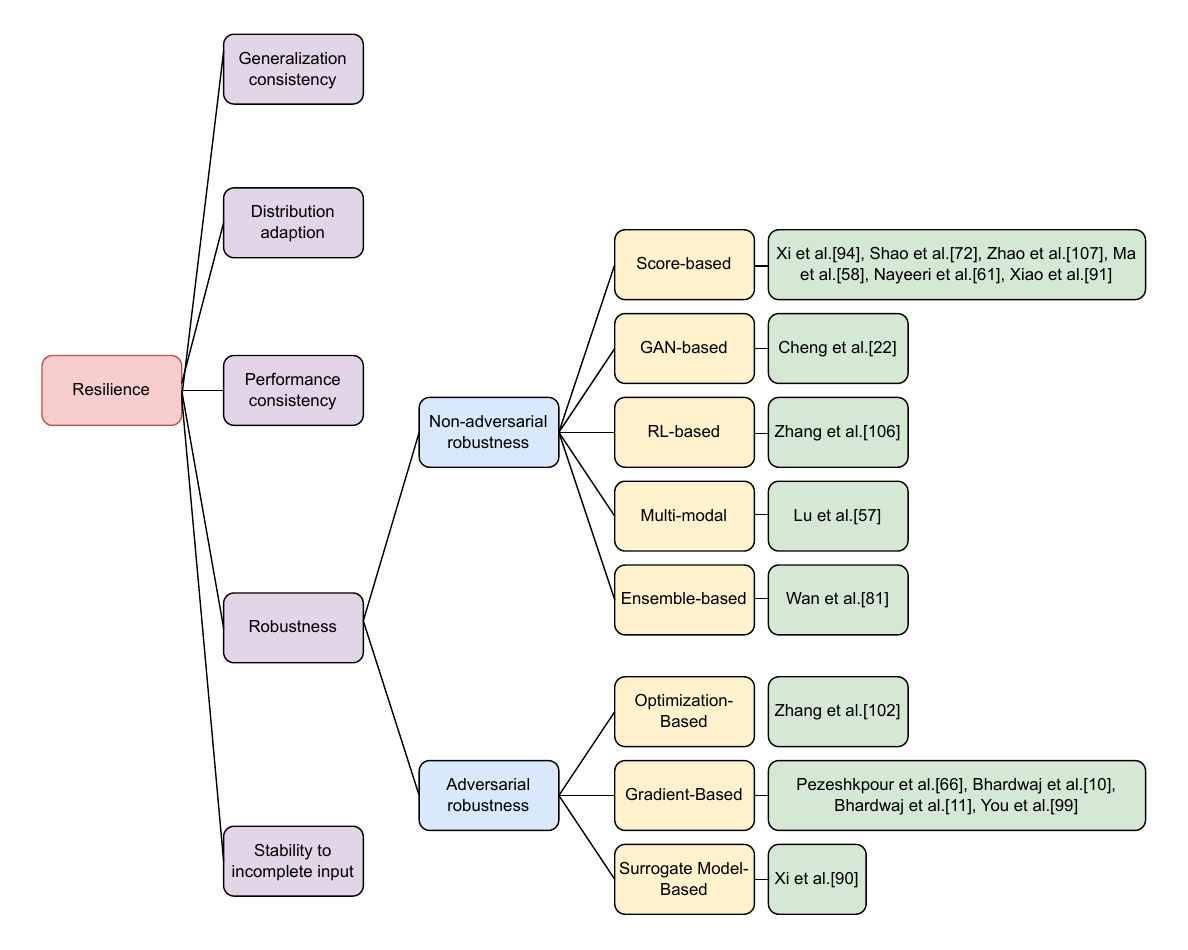}
    \caption{Categorisation of different works based on the underlying approaches in the context of resilience for knowledge graph embedding models}
    \label{fig:resiliencykg}
\end{figure*}

\section{Resilience in KGE}\label{sec:reslience-KGE}


The existing works considering resilience in KGE models mainly focus on a specific aspect, that is building KGE models that are resilient against the noise present in the KGs. To this end, there exist a number of such contributions~\cite{Xie0LL18, ShanBLJL18, ShaoLZXX21, NayyeriVSAYL21, ZhangZZLXHX23}. These works do not give any formal definition of resilience, rather consider, if the performance of the model does not degrade with noise present in the KG, then the underlying KGE model is said to be resilient. This, however, is not resilience based on the definition provided in Section~\ref{sec:resiliency-definition} where we defined resilience as a multi-faceted term that takes into account many aspects. Based on our definition of resilience, the work on resilience to safeguard against {\em noises} in the KGs mostly aligns with the definition of robustness, more specifically, the non-adversarial robustness (as defined in Equation 9). Therefore, we categorize this line of work related to resilience against noise as part of the non-adversarial robustness.

Moreover, there exist some works which concentrate on constructing resilient systems leveraging KGs~\cite{10371335,10021017,anjomshoaa2023data,dagnas2024resilience}. For instance, the works in~\cite{10371335} focus on building a KG-based risk assessment framework to improve the resiliency of supply chain management. A KG is built in~\cite{10021017} from the natural disaster data to improve the disaster management department's resilience towards such incidents. A similar sort of study is done in~\cite{anjomshoaa2023data} to employ a resilient management system in case a crisis happens in a city. To assess the resiliency of the cyber-physical system for a water management system, Dagnas et al.~\cite{dagnas2024resilience} utilized the KG as a modeling graph. 

Apart from the works mentioned earlier, there does not exist any other works that consider resilience as part of KGs or KGE models, therefore, in this work, we further survey the existing literature by considering individual aspects of resilience as described in Equation (1)--(10).  Figure~\ref{fig:resiliencykg} shows the categorization of the works found corresponding to the aspects of resilience that we described in Section~\ref{sec:resiliency-definition}. There exist works that focus on improving the generalizability of the KGE models~\cite{GuoWWWG17,LiCHSLC19,JainTG021,WuSCCLZW021}, or focusing on the logical consistency of the ontological rules~\cite{RocktaschelSR15,DemeesterRR16,GuoWWWG16,DuQWPLS17,DuQS18}, however, no work exists discussing generalization consistency, distribution adaption, performance consistency, and stability to incomplete input of KGE models.

More importantly, all the works found in regards to resilience in KGE can be distributed along two fields of robustness, namely adversarial and non-adversarial. There exists work such as by Zhu et al.~\cite{ZhuZZQSCH23} where they proposed to use KG to tackle the distribution shift problem for the {\em few-shot} learning approach. More specifically, by using KGs, the aim is to capture the semantic relationship between different categories of instances.  Despite the data samples originating from diverse distributions, they frequently possess shared auxiliary knowledge, along with prior semantic relationships between classes. For this, KG can serve to find out when such a distribution shift occurs and help the underlying model to adapt. 

Note that there exists a type of KG called {\em temporal} KG which dynamically evolves over time~\cite{ji2021survey}. The idea therein is to model the temporal information in KGs to keep track of how different assertions evolve over time. For this, the KG is defined as $\mathcal{G}_t := \{ (h, r, t, t') \in \entities \times \relations \times \entities \times \mathcal{T} \}$, where $t'$ basically points to the timestamp. For instance, `(Barack Obama, President\_of, USA, [2009-2017])' is an assertion associated with a timestamp for which it is true. Temporal KGs are called dynamic KGs since they are not static and evolve with the addition of new timestamps corresponding to assertions, and there are KGE models  that attempt to learn embeddings for such graphs, such as~\cite{AminVDN20,TangYLWYCS20,HeZLXW22,SunSZH23,YuGYXZLLY23}. Although such KGE models learn to map entities and relations into an embedding space which changes over time, we do not expect the learned embeddings to adapt to the distribution shift of the KG. This is because such KGE approaches assume that, alike typical KGs, the distribution of temporal KGs does not change and only a new timestamp is added by replacing the old timestamp, for instance, $(h, r, t, t')$ is replaced by $(h, r, t, t^*)$. The underlying KGE models therefore only need to adapt based on this newly added timestamp. A survey on temporal KGE models can be found here~\cite{abs-2308-02457}. 

Since there do not exist any works considering the aspects of resilience apart from robustness, in the following, we first describe the related works encompassing the two areas that exist regarding the resilience of KGE, namely adversarial and non-adversarial robustness, and thereafter we discuss possible future directions.

\section{Robustness}
\label{sec:robustness}
The concept of resilience in KGE models, as discussed earlier, encompasses multiple aspects. However, the existing literature primarily focuses on robustness, which can be divided into two main types: adversarial robustness and non-adversarial robustness.
Adversarial robustness concerns the model's ability to withstand intentional attacks, where malicious entities modify the knowledge graph (KG) to compromise the KGE model's performance. In contrast, non-adversarial robustness deals with the model's resilience against noise and inconsistencies naturally present in KGs, without any malicious intent.
In the following subsections, we provide a detailed survey of the current research related to these two areas of robustness in KGE models. We begin with adversarial robustness, followed by an exploration of non-adversarial robustness techniques.

\subsection{Adversarial robustness}

Despite its significance, the existing works on adversarial robustness in KGE models are in its infancy. The majority of studies focus on generating adversarial examples to deliberately manipulate the knowledge graph and assess the vulnerability of KGE models~\cite{ZhangZGMSL019,PezeshkpourTS19,BhardwajKCO20,BhardwajKCO21,YouSDZPYF23,XiDLPJLXM023}. These works proposed methods for {\em attacking} the KGE models by generating adversarial examples to study the robustness of the existing KGE models. Below we categorize them based on the approaches used to perform such attacks.

\subsubsection{Optimization Approach}
One of the earlier works in this area~\cite{ZhangZGMSL019} introduced a data poisoning attack strategy, aiming to alter the score of a target triple $(h_t, r_t, t_t)$ by modifying the KG. To achieve the poisoning goal, they assumed the attacker had a fixed budget (for instance, like $\Delta$ in Equation~\ref{equation:robustness}) in terms of the number of changes that could be done on the KG. To this end, they have given two attack strategies, namely direct and indirect attack. 

\begin{itemize}
    \item \textbf{Direct Attack}: The direct attack  involves identifying the necessary perturbations, such as adding or removing triples, to achieve the attacker's objective—for example, reducing the likelihood of a target fact $( h_t, r_t, t_t)$ being true. This process starts by determining the embedding shift $\epsilon$ required for either the head entity $\mathbf{h}_t$ or the tail entity $\mathbf{t}_t$ of the target triple to ensure that the new score $\phi'_\Theta( h_t, r_t, t_t )$, learned on the adversarially modified KG, is lower than the original score $\phi_\Theta(h_t, r_t, t_t)$. Potential perturbations are evaluated and ranked based on a scoring metric, guiding the selection of the most effective changes. The top $M$ perturbations are then chosen using an optimization technique, taking into account the attacker's budget and constraints.
    \item \textbf{Indirect Attack}: However, performing a direct attack that involves shifting the embeddings of the target triple might be detected by using some kind of sanity check. Hence, to make the attack stealthy, the authors in~\cite{ZhangZGMSL019} proposed indirect attack which involves shifting the embedding of the entities which are some $k$-hops away from the target triple $\triple{h_t}{r_t}{t_t}$. The changes would then propagate to the required embedding shifting of the target triple. 
\end{itemize}

While direct and indirect adversarial attacks on knowledge graph embeddings leverage optimization techniques to identify the most impactful perturbations, there are drawbacks when using such an  approach in this process. One key limitation is that KGs have a highly discrete and complex structure, making it difficult to navigate the search space effectively using straightforward optimization.

\subsubsection{Gradient-Based and Attribution Attacks}

Building on the ideas of direct and indirect attacks that target specific triples or entities in the knowledge graph, Pezeshkpour et al.~\cite{PezeshkpourTS19} followed a typical gradient-based approach to find out the most influential neighboring triple $\triple{h'_t}{r'_t}{t_t}$ of the target triple $\triple{h_t}{r_t}{t_t}$, the removal ($\kg \setminus \{\triple{h'_t}{r'_t}{t_t}\}$) or  addition ($\kg \cup \{\triple{h'_t}{r'_t}{t_t}\}$) of which would maximize the attack objective which can be defined as: 
\[\argmax_{(h'_t, r'_t)} \phi_\Theta \triple{h_t}{r_t}{t_t} - \phi'_{\Theta} \triple{h_t}{r_t}{t_t},\]

where $\phi'_{\Theta} \triple{h_t}{r_t}{t_t}$ defines the score when trained on either $\kg \setminus \{\triple{h'_t}{r'_t}{t_t}\}$ or $\kg \cup \{\triple{h'_t}{r'_t}{t_t}\}$. However, searching for such a $h'_t, r'_t$ is computationally expensive since the size of the search space is $|\entities| \times |\relations|$ (\emph{number of entities in $\kg$ $\times$ number of relations in $\kg$}), and moreover, the search space is discrete. Therefore, unlike the previous work~\cite{ZhangZGMSL019}, the authors herein modified the objective function by performing the search in the embedding domain, i.e., in the continuous space which gives the embedding for the optimal   head and relation as $\mathbf{h}'_t, \mathbf{r}'_t$. Thereafter, an autoencoder is used to get $h'_t, r'_t$ from $\mathbf{h}'_t, \mathbf{r}'_t$. However, one of the drawbacks of this approach is that it could only be used for multiplicative KGE models and moreover it does not take into account the nature of the KGE model being attacked.

Bhardwaj et al.~\cite{BhardwajKCO20} proposed a poisoning attack on KGE models by leveraging the inductive capabilities of these models, encapsulated through relationship patterns such as symmetry, inversion, and composition within a knowledge graph. Their approach aims to either decrease or increase the model's confidence in predicting a target triple $h_t, r_t, t_t$. For instance, if the attacker's goal is to decrease the score, they aim to ensure that $\phi_\Theta(h_t, r_t, t_t) > \phi'_\Theta(h, r, t )$, where $\phi'_\Theta$ is the model learned on the KG modified with the addition of adversarial triples, referred to as {\em decoy triples.}
These decoy triples are selected based on the inductive relation patterns that the KGE model captures. For example, if there exists a target triple $h, r, t$ composed of $h_t, r_1, \bar{t}$ and $\bar{t}, r_2, t_t$, an additive model that captures the symmetry relationship can be exploited, such that $\mathbf{r}_1 + \mathbf{r}_2 = \mathbf{r}$. The model then selects a relation $\mathbf{r}_t$ as the target relation, minimizing the Euclidean distance $|\mathbf{r}_t - (\mathbf{r}_1 + \mathbf{r}_2)|$. By doing so, the method identifies the relation that strongly captures the symmetry.
Once the target relation is chosen, two decoy triples are added in the form of $h, r_1, t^*$ and $t^*, r_2, t'$. These added triples manipulate the inductive properties of the KGE model, indirectly decreasing the score of the original target triple $h, r, t $. By exploiting the underlying inductive patterns that KGE models learn, such as symmetry and composition, this approach makes the target triple less likely to be predicted as true.

Bhardwaj et al.~\cite{BhardwajKCO21} further extended their approach by employing instance attribution methods from the domain of interpretable machine learning to carry out data poisoning attacks on KGE models. The aim of these attacks remains similar to their previous work: reducing the likelihood of the target triple $\triple{h_t}{r_t}{t_t}$ being correctly predicted by the KGE model. They specifically defined the attacker's capability as the ability to make a single change (either by removing or adding a triple) within the neighborhood of the target triple. The neighborhood is constructed based on triples that share either the subject or object of the target triple, formally defined as $\mathcal{H} = { (h_n, r_n, t_n) \mid h_n \in {h_t, t_t} \vee t_n \in {h_t, t_t} }$.
To identify which triple should be manipulated, they introduced an {\em influence score} $\mathcal{I}(\triple{h_t}{r_t}{t_t}, \triple{h}{r}{t})$. This score measures the effect that a particular training triple $\triple{h}{r}{t}$ has on the model's prediction for the target triple $\triple{h_t}{r_t}{t_t}$. A larger influence score indicates that removing the triple $\triple{h}{r}{t}$ would significantly reduce the predicted score for $\triple{h_t}{r_t}{t_t}$. However, directly retraining the KGE model for each triple removal is computationally expensive. To tackle this, the authors adopted techniques from interpretable machine learning, specifically using similarity metrics in the embedding space, to approximate the influence score.
They explored several gradient-based similarity metrics, including dot product, cosine, and $l_2$ similarity, to estimate the impact of removing a triple on the model's prediction of the target triple. Using the Hessian matrix, they further refined their influence score to capture the nuanced effects of removing a triple on the prediction. This approach enables them to efficiently identify the most influential triples and carry out the poisoning attack effectively.

You et al.~\cite{YouSDZPYF23} recently proposed a model-agnostic, semantic, and stealthy data poisoning attack on KGE models, addressing several aspects: black-box attack, semantically preserving poisoning, and stealthiness by ensuring good performance for clean triples. Unlike previous works, their approach focuses on inserting \textit{indicative paths} rather than individual triples to maximize the prediction probability of a target poisoned triple. The attack goal can be formalized as:
\[
\max_{\hat{T}} \phi_{\Theta}(h_t, r_t, t_t),
\]
where $\hat{T}$ is the set of triples in the indicative path.

In their approach, the key idea is to add \textit{indicative paths} that comprise more than one triple, which encourages the KGE model to predict the malicious fact as true. They translate the relation of the malicious fact into a sequence of relations using a \textit{path template}. For example, a path template $p_{h_t \rightarrow t_t}$ could be $h_t \xrightarrow{r_1} e \xrightarrow{r_2} t_t$, where $r_1, r_2$ is a relation template, and $e$ is an entity satisfying certain semantic constraints.
The steps involve using the Path Ranking Algorithm (PRA) to generate candidate relation paths. Next, they leverage \textit{semantic constraints} by selecting entities for the indicative paths that adhere to the domain and range constraints of the relations involved. The selection is carried out using a gradient-based search technique to find the indicative paths that maximize the prediction score for the target triple $h_t, r_t, t_t$. By ensuring that the added paths align with semantic constraints and maximize the plausibility of the malicious triple, their approach not only remains stealthy but also effectively biases the model's predictions towards the attacker's objective. This method is validated through extensive evaluations on benchmark datasets, demonstrating its effectiveness in achieving a high attack success rate under various opaque-box settings.

\subsubsection{Surrogate Model-based Attack}
Xi et al.~\cite{XiDLPJLXM023} introduced ROAR, an attack strategy designed to attack KGE models through both knowledge graph poisoning and \textit{query misguiding}. ROAR particularly focuses on downstream applications where KGEs provide answers to user queries. The goal of the attack is to manipulate the response to a specific query by poisoning the knowledge graph in a manner that maximizes the probability of the targeted fact being true.
The attack begins by generating a \textit{surrogate knowledge graph} $\mathcal{G'}$ from the original one. This surrogate graph is used to build a \textit{surrogate knowledge graph reasoner}, which consists of a surrogate embedding function $\phi'$ and a transformation function $\psi$. These functions are trained on a set of question-answer pairs sampled from $\mathcal{G'}$. The challenge here is that directly searching for poisoning facts that make the targeted fact true in the discrete space of the knowledge graph is computationally expensive.

To overcome this, the authors first employ \textit{latent space optimization}. They search for an \textit{anchor entity} connected to the target fact and identify facts in the embedding space which, when added, increase the probability of the targeted fact. These potential additions to the graph are gathered in a set of embeddings $\{\mathbf{h}_i, \mathbf{r}_i, \mathbf{t}_i\}_{i=1}^N$. 
Next, the effectiveness of adding each potential fact is assessed using a \textit{fitness score}, which indicates how much each fact's addition would increase the plausibility of the target fact. Based on this score, the top $n_g$ facts are selected for addition to the knowledge graph. This selection process ensures that only the most influential facts are included in the poisoning attack, thereby maximizing the impact on the targeted queries.
This two-step process of latent space optimization followed by fitness-based selection makes ROAR a highly adaptable and effective adversarial attack against KGEs, especially in scenarios where downstream applications rely on the knowledge graph for query resolution.

\subsection{Non-adversarial robustness}

While adversarial robustness focuses on defending against malicious attacks, non-adversarial robustness concerns the model's resilience to naturally occurring noise and inconsistencies in KGs. Real-world KGs are often incomplete, contain errors, and exhibit conflicting information due to the diverse sources from which they are constructed. A robust KGE model should be able to handle these imperfections without significantly compromising its performance. Several approaches have been proposed to improve the robustness of KGE models under noisy KGs, ranging from confidence score-based methods to GAN-based frameworks, reinforcement learning techniques, multi-modal approach, ensemble approach~\cite{Xie0LL18,ShanBLJL18,ZhaoFG19,Cheng2020NoiGANNA,ShaoLZXX21,NayyeriVSAYL21, MaZWGHQW22,LuWJHL22,WanDPW20,ZhangZZLXHX23}. Below we discuss the existing works considering these approaches. 

\subsubsection{Confidence Score-based Approaches}

Confidence score-based approaches have been proposed to enhance the robustness of KGE models by quantifying the reliability of each triple within the KG. These methods assign a confidence score, trustworthiness value, or distance-based measure to each triple, allowing the model to prioritize more reliable data during training. The confidence scores guide the learning process, helping the model to distinguish between correct and noisy triples, thus reducing the impact of inaccuracies present in real-world KGs. In this section, we discuss several works that introduce different mechanisms for computing and utilizing confidence scores to improve the robustness of KGE models. These mechanisms range from local and global confidence scores to trustworthiness evaluations and distance-based assessments.

\paragraph{Local \& Global Confidence Score}

Xie et al.~\cite{Xie0LL18} introduced one of the earliest methods to address noise in knowledge graphs by developing KGE models that are robust to such noise. They proposed a novel approach known as the confidence-aware knowledge representation learning (CKRL) framework, which assigns a confidence score to each triple in the KG. This score indicates the correctness and significance of each triple, allowing the model to prioritize more reliable triples during learning.
Their model builds upon the translation-based KGE approach, specifically utilizing TransE~\cite{BordesUGWY13}, as the scoring function $\phi_{\Theta}$. The standard margin-based ranking loss function~\cite{ChenLLML09} was modified to incorporate the confidence scores of triples. The revised objective function aims to minimize the impact of noisy triples by giving higher importance to more reliable triples. Specifically, they introduced the confidence-aware loss function:

\[
\sum_{(h,r,t) \in S^+} \sum_{(h,r,x) \in S^-} [\gamma + \phi_{\Theta}(h,r,t) - \phi_{\Theta}(h,r,x)] \cdot C(h,r,t),
\]

where $\gamma$ is the margin, and $S^+$, $S^-$ are the sets of positive and negative triples, respectively. Here, $C(h,r,t)$ is the confidence score for the triple $h, r, t$. A higher confidence score signals that the model should prioritize this triple during training. In essence, triples with higher scores are treated as positive, while those with lower scores are treated as negative.

The computation of the confidence score $C(h,r,t)$ involves two components: local and global confidence scores. 

\begin{itemize}
    \item \textbf{Local Confidence Score:} This score evaluates how well a triple conforms to the translation assumption within the KGE model. The triple's quality is updated iteratively during training. If a triple does not align with the translation rule, its confidence decreases by a geometric rate $\alpha$. Conversely, if it does align, the confidence increases at a constant rate $\beta$. This iterative adjustment ensures that the confidence scores reflect the quality of triples over time.
    
    \item \textbf{Global Confidence Score:} To provide a more holistic assessment, Xie et al. considered multi-step relation paths between entities, which inform the global confidence score. The intuition is that the existence of reliable paths in the KG, connecting the head and tail entities, provides additional evidence supporting the correctness of a triple. Therefore, both the local and global scores are combined to generate the final confidence score $C(h,r,t)$.
\end{itemize}

By combining these scores, CKRL effectively learns embeddings while simultaneously detecting and mitigating the influence of noise in the KG. This pioneering work laid the foundation for later developments in confidence-aware KGE models.

In a later work, Shan et al.~\cite{ShanBLJL18} argued that the confidence score mechanism proposed by Xie et al.~\cite{Xie0LL18} could lead to the ``zero loss problem.'' This issue occurs when the negative triples sampled during training quickly fall outside the margin in the ranking loss function, resulting in zero loss. When this happens, the negative triples cease to contribute to refining the model's embeddings, leading to slow convergence, reduced accuracy, and diminished effectiveness in detecting noise within the knowledge graph.
To address this problem, Shan et al. introduced a novel \textit{confidence-aware negative sampling method}. They proposed a mechanism to assign a confidence score not just to positive triples but also to negative triples, with the goal of identifying high-quality negative triples that could contribute more significantly to the model's learning process.
The key idea is to incorporate the confidence scores of negative triples into the training process. Shan et al. defined the confidence of a negative triple $h', r, t'$ based on a softmax function over a set of candidate negative triples $\mathcal{N}(h, r, t)$:

\[
C(h', r, t') = \frac{\exp(\phi_\Theta(h', r, t'))}{\sum_{h'', r, t'' \in \mathcal{N}(h, r, t)} \exp(\phi_\Theta(h'', r, t''))}.
\]

Here, $\phi_\theta(h', r, t')$ is the score function for the negative triple $(h', r, t') $, and $\mathcal{N}(h, r, t)$ is the set of candidate negative triples. The softmax function normalizes the scores to provide a probability distribution over the negative triples, allowing the model to select the negative triple with the highest confidence, which has a more significant impact on refining the embeddings during training.
Shan et al. also proposed a modification to the marginal ranking loss function introduced in~\cite{Xie0LL18}. Rather than focusing on the quality of a single negative triple, they consider the expected score of a group of candidate negative triples. The revised loss function is defined as:

\[
\mathcal{L} = \sum_{h, r, t  \in S^+} \left[ \gamma + \phi_\Theta(h, r, t) - \mathbb{E}_{h', r, t' \sim \mathcal{N}(h, r, t)} \left[ \phi_\Theta(h', r, t') \right] \right]_+,
\]

where $\gamma$ is the margin, $S^+$ is the set of positive triples, and $\mathbb{E}_{h', r, t'  \sim \mathcal{N}(h, r, t)}[\phi_\Theta(h', r, t')]$ is the expected score of the sampled negative triples from $\mathcal{N}(h, r, t)$. The expectation ensures that the model considers a diverse set of high-quality negative triples during training, which prevents the zero loss problem by keeping the negative samples within the margin.
By using the confidence-aware negative sampling method and the modified loss function, Shan et al.'s approach dynamically adjusts the sampling of negative triples. This method alleviates the zero loss problem by actively incorporating high-quality negative triples into the learning process, which lie closer to the decision boundary defined by the margin. Consequently, the model becomes more robust and better at distinguishing noisy triples.

Shao et al.~\cite{ShaoLZXX21} extended the confidence score-based methods by introducing a novel framework called DSKRL (Dissimilarity-Support-Aware Knowledge Representation Learning) to handle noise in KGs more effectively. Their approach incorporates two main components: triple dissimilarity and triple support, leveraging both structural and auxiliary information in KGs.
While the former measures how well the entities and relations in a triple match, using entity hierarchical types and relation paths, the latter combines local and dynamic path support to assess a triple's credibility. Triple dissimilarity score is computed using entity hierarchical type Information ($E_{HT}(T_h, r, T_t)$) and using the relational path information ($R_P(h, P, t)$) as follows.
    \[
    E_{HT}(T_h, r, T_t) = \| T_h + r - T_t \|_2,
    \]
    where \( T_h \) and \( T_t \) are the hierarchical type embeddings for the head and tail entities, and \( r \) is the relation embedding.
    \[
    R_P(h, P, t) = \frac{1}{Z} \sum_{p \in P(h, t)} R(p \mid h, t) E(h, p, t),
    \]
    where \( R(p \mid h, t) \) is the path reliability and \( Z \) is a normalization factor. Finally, the overall triple dissimilarity is given by,
\[
PT(h, r, t) = E_{HT}(h, r, t) + R_P(h, P, t).
\]

The triple support measures the credibility of the matching extent for the triples. It combines local triple support ($LS(h, r, t)$) and dynamic path support ($DPS(h, r, t)$) to make this estimation more reliable.

    \[
    LS(h, r, t) = \gamma LS(h, r, t), \quad \text{if } Q(h, r, t) \leq 0.
    \]

  To further utilize the relation path information, they introduce dynamic path support as follows.
    \[
    DPS(h, r, t) = \sigma \left( \sum_{p_i \in S(h, t)} \frac{R(p_i \mid h, t)}{Q_{DPS}(r, p_i)} \right),
    \]
    where \( Q_{DPS}(r, p_i) \) measures the path quality.
Finally, the local and dynamic support scores are combined as follows.
\[
S(h, r, t) = \lambda_1 LS(h, r, t) + \lambda_2 DPS(h, r, t).
\]

After computing both the dissimilarity estimator and triple support they combine them to improve the noise resilience in KGE models.

\paragraph{Trustworthiness Score}

In~\cite{ZhaoFG19}, Zhao et al. proposed a method to compute the \textit{trustworthiness value} of a triple based on entity type instances and entity descriptions, leveraging the semantic information in knowledge graphs. The underlying idea is that certain entity types inherently have varying credibility when filling slots in triples. For instance, a living entity (e.g., \texttt{/people/person}) is generally more suitable than a non-living one (e.g., \texttt{/book/written\_work}) for the triple (?, \texttt{was\_born\_in}, \texttt{the State of Hawaii}). This intuition helps assign trustworthiness scores to triples.
To quantify this, Zhao et al. introduce a model called \textbf{TransT} that calculates the trustworthiness of triples using two main components, Entity Type Trustiness (TT) and Entity Description Trustiness (DT). These components are then integrated into an overall trustiness score that informs the embedding learning process.
The TT score based on entity types, \( TT(h, r, t) \), captures the compatibility of entity types with a given relation. It is defined as follows.
\[
TT(h, r, t) = \frac{1}{Z} \sum_{(h_i, t_i) \in \mathcal{T}(r)} \exp(-d(h_i, r, t_i)),
\]
where \( \mathcal{T}(r) \) is the set of type pairs \( (h_i, t_i) \) corresponding to the head and tail types of entities for relation \( r \), \( d(h_i, r, t_i) \) is a distance function measuring how well the head type \( h_i \) and tail type \( t_i \) align with the relation \( r \), and \( Z \) is a normalization factor.
The intuition is that entities belonging to more trustworthy types for the given relation will have higher compatibility, resulting in a higher trustiness score.
To further enhance trustworthiness estimation, Zhao et al. incorporate the entity descriptions into the computation. This component considers the semantic content of entity descriptions to assess the credibility of the triple. The description-based trustiness, \( DT(h, r, t) \), is calculated using the cosine similarity between the vector representations of the entity descriptions and the relation:
\[
DT(h, r, t) = \cos(\mathbf{d}_h + \mathbf{r}, \mathbf{d}_t),
\]
where \( \mathbf{d}_h \) and \( \mathbf{d}_t \) are the vector representations of the descriptions of the head and tail entities, respectively, and \( \mathbf{r} \) is the vector for the relation \( r \). The cosine similarity measures the alignment between the combined description vectors and the relation vector.
The final trustworthiness score of a triple \( (h, r, t) \) is a weighted combination of the type-based and description-based trustiness scores:
\[
T(h, r, t) = \alpha \cdot TT(h, r, t) + \beta \cdot DT(h, r, t),
\]
where \( \alpha \) and \( \beta \) are hyperparameters that control the relative contributions of the type and description components.
The computed trustworthiness score \( T(h, r, t) \) is then integrated into the training of the knowledge graph embedding model. The model assigns higher importance to triples with higher trustworthiness, thereby enhancing the robustness of the learned embeddings against noisy or incorrect triples. The final loss function for the model is modified as:
\[
\mathcal{L} = \sum_{(h, r, t) \in S^+} T(h, r, t) \cdot \left[\gamma + \phi_{\Theta}(h, r, t) - \phi_{\Theta}(h', r, t')\right]_+,
\]
where \( S^+ \) is the set of positive triples, \( \gamma \) is the margin. The trustworthiness score \( T(h, r, t) \) acts as a weight, prioritizing the learning of more trustworthy triples.
This approach effectively improves the embeddings by ensuring that the model focuses on more reliable information within the KG.

Ma et al.~\cite{MaZWGHQW22} proposed a noise detection method called PTrustE, which integrates path trustworthiness and triple embeddings to address noise in KGs. Unlike traditional methods that focus on either local or global properties, PTrustE combines both to detect and categorize noisy triples effectively. Given a triple $(h, r, t )$, PTrustE first searches all paths between the head entity $h$ and the tail entity $t$. Each path consists of a series of intermediate entities and relations, which are then used to compute both local and global trustworthiness scores. Specifically, two types of trustworthiness are introduced. Local Triple Trustworthiness (LTT) and Global Triple Trustworthiness (GTT).

To detect noise in a KG, first of all, PTrustE constructs a correlation-based path trustworthiness network. This network learns the global and local features of the paths from the head entity $h$ to the tail entity $t$ of a triple. The path is initialized as a sequence of nodes, where each node represents a triple. Each triple $(x_i, r_i, x_{i+1})$ in the path is embedded into a continuous vector space, denoted as $N'_i = (x'_i, r'_i, x'_{i+1})$.
To measure the trustworthiness of a triple, PTrustE employs a Correlation-Based Probability Logic Network (CPLN). This network captures the correlation between entities and relations using both logical rules and probabilistic scoring:
\[
\varphi(i, j) = \sum_{n=1}^{4} \varphi_n(i, j),
\]
where $\varphi(i, j)$ is the correlation score between triples $f_i$ and $f_j$. Each $\varphi_n(i, j)$ is computed based on logical operations that consider the similarities between entities and relations in the triples. For example, $\varphi_1(i, j)$ considers the correlation between head entities and tail entities and whether their relations are positively correlated.
Afterwards, GTT and LTT scores are computed. GTT is defined as the product of the relative distances between triples along the path,
$\text{GTT}(i, j) = N'_i \cdot (N'_j)^\top \cdot W_0$,
where $W_0$ is a parameter matrix learned during training. LTT on the other hand captures the correlation among entities and relations of adjacent triples in the path. It can be computed as follows.
\[
\text{LTT}(i, i+1) = \sum_{n=1}^{4} \varphi_n(i, i+1),
\]
where each $\varphi_n$ uses logical operations (AND, OR) to combine the correlations between entities and relations.
PTrustE combines the GTT and LTT to obtain the node's trustworthiness score as, $G_i = \text{GTT}(i, i+1) \cdot \text{LTT}(i, i+1).$
Next, these node scores are input into a Bi-directional Gated Recurrent Unit (BiGRU) to capture the sequential nature of the path. The final path score matrix is computed using a linear transformation and a softmax function, $M(P) = \text{softmax}[\text{linear}[\text{concat}(G^{(1)}_{m-1}, G^{(2)}_{m-1})]]$,
where $M(P)$ represents the path score matrix. The final path trustworthiness score ($S(f)$) is defined as the L2-norm of the path score matrix as follows.
\[
S(f) = \max \left[ \| M(P) \|_2 \right],
\]
where the path with the maximum trustworthiness score is used for triple representation learning. Through extensive experiments, PTrustE demonstrates superior performance in noise detection, showing that integrating path features with logical reasoning leads to a more robust KG embedding model.

\paragraph{Distance-based Confidence Score}

Nayyeri et al.~\cite{NayyeriVSAYL21} introduced a modification to the marginal ranking loss function to handle noisy data in knowledge graphs (KGs), particularly focusing on incorrect triples. Their approach does not build on the previous confidence score-based works but instead introduces a distance-based strategy to identify and manage noisy triples effectively.
In their method, the authors define separate objective functions for positive and negative triples and then combine them into a unified loss function. One key component of their approach is a distance function, $\tau_{h,t}^r: S^+ \rightarrow [0, \infty]$, which intuitively measures the likelihood of a triple being correct or noisy. During the optimization process, this distance is constrained to lie within the range $[0, \gamma]$, where $\gamma$ serves as a discriminator that separates positive and negative triples.
A probability function, $P(\tau_{h,t}^r)$, is employed to assign a score based on the computed distance. A high probability indicates a high likelihood of the triple being incorrect (noisy), whereas a lower probability suggests a higher confidence in the triple's correctness. The objective is to minimize the overall loss by maximizing the likelihood of correct triples and minimizing the likelihood of noisy ones. This can be formally expressed as:
\[
\min_{(h, r, t) \in S^+, \tau_{h,t}^r} \prod P(\tau_{h,t}^r),
\]
subject to the constraint:
$\mathcal{Q}(\gamma - \phi_{\Theta}(h, r, t)) \geq \mathcal{Q}(\tau_{h,t}^r)$,
where $\mathcal{Q}$ represents a ranking function.
The constraint ensures that the distance $\tau_{h,t}^r$ does not exceed the margin $\gamma$, thereby maintaining the positive examples on one side of $\gamma$. For negative examples $(h', r, t')$, the constraint is modified to $
\mathcal{Q}(\phi_{\Theta}(h', r, t') - \gamma) \geq \mathcal{Q}(\tau_{h',t'}^r).$
The final loss function integrates both positive and negative triples, formulated as:
\[
\min_{\{(h, r, t)\} \in S^+, \{(h', r, t')\} \in S^-, \tau_{h,t}^r, \tau_{h',t'}^r} 
\sum_{(h, r, t) \in S^+} \log P(\tau_{h,t}^r) + \sum_{(h', r, t') \in S^-} \log P(\tau_{h',t'}^r),
\]
subject to the conditions: $
\mathcal{Q}(\gamma - \phi_{\Theta}(h, r, t)) \geq \mathcal{Q}(\tau_{h,t}^r), \quad \mathcal{Q}(\phi_{\Theta}(h', r, t') - \gamma) \geq \mathcal{Q}(\tau_{h',t'}^r).
$

The core idea is that if a triple $( h', r, t' )$ is incorrectly identified as a false negative but should belong to the positive set based on inference rules, the distance $\tau_{h',t'}^r$ will be close to zero. The proposed loss function thus assigns a lower score to such triples, reflecting the uncertainty associated with their classification.

\subsubsection{GAN-based Approach}

NoiGAN~\cite{Cheng2020NoiGANNA} extends the idea of confidence score proposed in~\cite{Xie0LL18}. They argued, similar to the previously described approaches, that using only the confidence score as an indication of how well a triple fits to the KGE model might lead to bias and uncertainty. Therefore, the confidence score $C(h,r,t)$ in this work is learned by using a {\em generator} and {\em discriminator} as a generative adversarial network (GAN). More specifically, they proposed a learning framework inspired by the adversarial training~\cite{MiyatoMKI19,BalunovicV20,LiPCZLL23} methods. 
In the GAN framework, NoiGAN consists of two main components: a generator and a discriminator. The generator is designed to generate noisy triples, while the discriminator is trained to distinguish between true and noisy triples, ultimately computing the confidence score for each triple. During training, the KGE model uses this confidence score as a guiding signal to eliminate noisy data, resulting in noise-aware embeddings. 

Given a true triple $(h, r, t)$, the generator generates a noisy triple $(h', r, t')$ from an initially generated negative sample candidate set $\mathcal{N}(h,r,t)$. This is achieved through a neural network that takes as input the embedding vectors of the triple $(h', r, t')$ and outputs a probability indicating the plausibility of the triple being noisy. More formally, the generator aims to maximize the expected reward:
\[
R_G = \sum_{(h,r,t)} \mathbb{E}_{(h',r,t') \sim G(\cdot | (h, r, t); \Theta_G)} [\log f_D(h', r, t')],
\]
where $f_D(h', r, t')$ is the probability predicted by the discriminator that the generated triple $(h', r, t')$ is true. The generator uses reinforcement learning to generate triples that can effectively fool the discriminator. The discriminator, on the other hand, acts as a noisy triple classifier. It aims to distinguish between true triples and noisy triples generated by the generator. Given the difficulty in obtaining labeled noisy triples in real-world scenarios, NoiGAN relies on the generator to produce noise samples. The discriminator's objective function is formulated as minimizing the cross-entropy loss:
\[
L_D = -\sum_{(h,r,t) \in T_C} \log f_D(h, r, t) - \sum_{(h',r,t') \in G(\cdot | (h, r, t); \Theta_G)} \log(1 - f_D(h', r, t')),
\]
where $T_C$ is the set of triples that are considered correct based on their confidence scores.
The noise-aware KGE model uses these confidence scores as a guiding mechanism to mitigate the impact of noise during training. This approach allows NoiGAN to mutually enhance both noise detection and KGE model learning through adversarial training.

\subsubsection{Reinforcement Learning Approach}

A recent work by Zhang et al.~\cite{ZhangZZLXHX23} proposes a multi-task reinforcement learning (RL) framework to make the KGE models robust by identifying and removing noisy triples from the training dataset. Unlike previous approaches that directly train on noisy datasets, this method first cleans the dataset before the training process, ensuring that the KGE models are learned on a noise-free graph. The authors define the state, action, reward, and the objective of the RL framework in the following manner.

\begin{description}

 \item[State.] Each state in the RL is represented as the set of triples that have already been selected as clean and the current triple that is under consideration. Mathematically, the state at time step $t$ can be defined as, $s_t = (T_{\text{selected}}, (h, r, t ))$, where $T_{\text{selected}}$ is the set of triples that have already been marked as clean up to time $t$, and $(h, r, t)$ is the triple being evaluated.

\item[Action] The RL agent takes an action to either select or reject the triple $(h, r, t)$. The action space $A$ consists of binary decisions, $A = \{0, 1\}$, where $1$ indicates selecting the triple as clean, and $0$ indicates rejecting it.

\item[Reward] The reward function is designed based on the scoring functions of multiple KGE models like TransE, DistMult, ConvE, or RotatE, along with a heuristic term that encourages the model to select more triples. The reward $R$ for a set of selected triples $T_{\text{selected}}$ is calculated as:
   \[
   R = \frac{1}{|T_{\text{selected}}|} \sum_{(h, r, t) \in T_{\text{selected}}} \phi_{\Theta}(h, r, t) + \alpha \frac{|T_{\text{selected}}|}{|T_{\text{total}}|},
   \]
   where $\alpha$ is a hyperparameter, and $|T_{\text{total}}|$ is the total number of triples in the KG.

\item[Objective]  The aim of the RL model is to maximize the expected reward by selecting those triples that exhibit higher plausibility. This is formalized as:
   \[
   \max_{Q} \mathbb{E}_{a \sim p_Q(a|s)} [R] + \lambda_1 \|u_c\|^2 + \lambda_2 \|v_r\|^2,
   \]
   where $Q$ represents the parameters of the RL model, $u_c$ is the common parameter shared among related relations for knowledge sharing, $v_r$ is the relation-specific parameter, and $\lambda_1, \lambda_2$ are regularization coefficients.

\end{description}

The authors highlight that this approach has the potential drawback of filtering out a large number of triples, which could include some correct triples. However, the RL framework's use of scoring functions from different KGE models helps mitigate this by making decisions based on the inferred relationships and plausibility scores.

\subsubsection{Multi-modal Knowledge Representation}
The work closest to the idea of robustness of KGE models-- as defined in Equation 9 -- is done by Lu et al.~\cite{LuWJHL22} where they propose multi-modal knowledge representation learning (MMKRL) to generate robust KGE models. The idea therein is to use several knowledge such as textual knowledge, entity description, visual knowledge to generate the embedding~\cite{TangCCW19,wang2019multimodal,xie2019integrating}.
MMKRL essentially consists of two main modules: knowledge reconstruction and adversarial training (AT) where the knowledge reconstruction module aligns and integrates various knowledge embeddings to reconstruct multi-modal knowledge graphs, while the AT module enhances robustness and performance using adversarial strategies. Consider  $\mathbf{h}_s, \mathbf{r}_s, \mathbf{t}_s$ as the structural embedding and  $\mathbf{h}_m, \mathbf{t}_m$ as the multi-modal embedding (coming from textual and visual embedding). To compute the plausibility of each assertions, a feedforward neural network to encode the input embeddings as, 
\[ \mathbf{h}'_s = f(W_s \mathbf{h}_s + b_h), \mathbf{r}'_s = f(W_s \mathbf{r}_s + b_r), \mathbf{t}'_s = f(W_s \mathbf{t}_s + b_t) \: \: \: \text{and}\]
\[ \mathbf{h}'_m = f(W_s \mathbf{h}_m + b^m_h), \mathbf{t}'_m = f(W_m \mathbf{t}_m + b^m_t) \]

Based on the TransE model, a multi-modal plausibility function is defined as, 
\[ \phi_{\Theta} = || \mathbf{h}'_s + \mathbf{r}'_s - \mathbf{t}'_s ||  + || \mathbf{h}'_m + \mathbf{r}'_s - \mathbf{t}'_m || + (|| \mathbf{h}'_m + \mathbf{r}'_s - \mathbf{t}'_s ||  + || \mathbf{h}'_s + \mathbf{r}'_s - \mathbf{t}'_m ||)\]

A marginal-based ranking loss is then used to generate the optimal embedding. However, to make the multi-modal embedding model more robust an adversarial training approach is also included where the embeddings are perturbed during the training process. The authors claimed that this step would make the model more robust.

\subsubsection{Ensemble Approach}

Wan et al.~\cite{WanDPW20} proposed an ensemble-based approach to enhance the robustness of the KGE models. Their method involves generating a set of diverse subgraphs from a given KG $\mathcal{G}$ and training an individual base learner for each subgraph. Due to the complexity of KGs, traditional graph sampling methods are not directly applicable. To address this, Wan et al. employ a random walk-based approach~\cite{lovasz1993random} to sample meaningful subgraphs.
The random walk process starts by selecting an initial fact $(h, r, t)$ uniformly at random from the KG $\mathcal{G}$. Then, the random walk samples a neighbor of the current node, following the relations in the KG. This sampling continues until a predefined boundary condition $L$ (e.g., a maximum path length or number of nodes) is met. After executing multiple random walks, a set of subgraphs $\{\mathcal{G}_1, \mathcal{G}_2, \dots, \mathcal{G}_n\}$ is generated.
For each subgraph $\mathcal{G}_i$, a shallow KGE model $\phi_{\Theta_i}$ is trained independently to obtain entity and relation embeddings. The model's goal is to learn an embedding function $\phi_{\Theta_i}(h, r, t)$ that maximizes the plausibility of triples in the subgraph. 
The final ensemble model combines the outputs of these $n$ base learners. Let $\phi_{\text{ensemble}}$ represent the final embedding function, which is defined as a weighted combination of the individual models:
\[
\phi_{\text{ensemble}}(h, r, t) = \sum_{i=1}^{n} \alpha_i \phi_{\Theta_i}(h, r, t),
\]
where $\alpha_i$ is the weight assigned to model $\phi_{\Theta_i}$ based on its prediction performance. The weights $\alpha_i$ are determined by an uncertainty measure, which reflects the predictive capability of each model on its corresponding subgraph. For example, the uncertainty can be calculated using entropy or variance in the predictions.
The robustness of this ensemble approach is then evaluated by injecting noise into the KG $\mathcal{G}$. Wan et al. demonstrate that their ensemble model performs significantly better than individual KGE models in the presence of noisy triples. This suggests that the ensemble effectively mitigates the impact of noise, thereby enhancing the resilience of KGE models.

\subsubsection{Robustness Evaluation}

Xiao et al.~\cite{Xiao00KMA24} address the problem of evaluating the robustness of Outstanding Facts (OFs) derived from Knowledge Graphs (KGs). An OF is defined as a statement highlighting how an entity stands out based on specific attributes when compared to its peers. OFs are extensively used in data journalism, fact-checking, and recommendation systems to provide insightful and striking information. However, Xiao et al. argue that such statements may be misleading if the underlying context is not fully considered. This connects to the earlier discussions on robustness in KGs, emphasizing that resilience is not only about handling adversarial noise but also about maintaining meaningfulness under contextual changes. Consider a KG $\mathcal{G}$ containing information about universities and their employees, including attributes like gender. An OF from this KG might state: ``At the American Council on Education (ACE), only 31\% of the employees are male.'' This statement could suggest a notable gender disparity at ACE-affiliated institutions. However, the robustness of this fact needs to be evaluated by considering the broader context and possible data variations.
To formalize this, Xiao et al. introduce the concept of robustness by analyzing how the ``strikingness'' of an OF changes under various perturbations. Let $\mathcal{S}(f)$ denote the strikingness of an OF $f$ in a given context. The goal is to ensure that $\mathcal{S}(f)$ remains consistent even when the context or data changes slightly. The authors propose two types of perturbations to evaluate this robustness:

\begin{compactenum}
    
\item[Entity perturbation] It assesses the robustness of an OF by replacing its context entity $c$ with a similar entity $c'$. Formally, let $c$ represent the context entity in the OF $f$. We replace $c$ with $c'$, where $c'$ is chosen based on its similarity to $c$. The similarity between entities $c$ and $c'$ is computed using a node similarity score in the KG. For example, it can be measured using the Jaccard similarity between the neighbor sets of $c$ and $c'$ in the KG:
\[
\text{Sim}(c, c') = \frac{|N(c) \cap N(c')|}{|N(c) \cup N(c')|},
\]
where $N(c)$ and $N(c')$ are the sets of neighbors of $c$ and $c'$, respectively. 

In the case of our example, replacing "ACE" with a similar entity, such as "American Association of State Colleges and Universities (AASCU)," leads to a new OF: "At AASCU, 50\% of the employees are male." If this new fact appears less striking (e.g., 50\% being closer to gender parity), it implies that the original OF's strikingness was context-dependent and, thus, not robust.

\item[Data perturbation] It  involves modifying the KG by adding or altering edges, thereby changing the peer entity set of the OF. For example, suppose new data is added, introducing another university affiliated with ACE that has a 70\% male employee ratio. This alters the OF to: "At ACE, 40\% of the employees are male." The addition of new data provides a more balanced view, reducing the perceived strikingness of the original OF. 

Formally, the relevance of a data perturbation is quantified using a head-tail relevance function, which measures the semantic connection of the newly added edges to the original fact. Given an added edge $(h', r', t')$, the head-tail relevance function $\mathcal{R}(h', r', t')$ evaluates whether the modification preserves the context's semantic integrity.

\end{compactenum}

The robustness of an OF is then defined by the expected strikingness $\mathbb{E}_{p(\mathcal{P})}[\mathcal{S}(f)]$ over a perturbation relevance distribution (PRD) $p(\mathcal{P})$. Mathematically, it can be expressed as:
\[
\mathbb{E}_{p(\mathcal{P})}[\mathcal{S}(f)] = \int_{\mathcal{P}} \mathcal{S}(f') \, p(\mathcal{P}) \, d\mathcal{P},
\]
where $\mathcal{P}$ represents a set of possible perturbations (both entity and data perturbations), and $f'$ is the resulting fact after applying perturbation $\mathcal{P}$. The OF is considered robust if its strikingness $\mathcal{S}(f)$ does not significantly deviate from the expected value $\mathbb{E}_{p(\mathcal{P})}[\mathcal{S}(f)]$ across different perturbations.
This method of evaluating robustness relates to earlier discussions in the literature about the importance of contextual integrity in KGE models. Much like ensuring that KGE models are resilient against adversarial attacks and noise (e.g., as described in works like Xie et al.~\cite{Xie0LL18} and Shan et al.~\cite{ShanBLJL18}), evaluating OFs for robustness ensures that their interpretations remain valid across different contextual scenarios. By performing these perturbation analyses, stakeholders can avoid jumping to conclusions based on OFs that may appear striking only within a narrow or specific context.

\subsubsection{KG-based Approaches Enhancing Robustness}
Note that, similar to using KGs to improve the resilience of several systems, there also exist a number of works that use KGs~\cite{YangHXL22,radtke2023increasing} and KGE model~\cite{LangBV22} to improve the robustness of ML models. However, the notion of robustness therein pertains to the effectiveness of performing the underlying tasks.

For instance, Lang et al.~\cite{LangBV22} propose the use of Knowledge Graph Embeddings (KGE) to develop more robust multi-object detection models. The main idea is to use KGEs to incorporate semantic knowledge into object detection, aiming to achieve more structured and semantically grounded predictions. Traditional object detection models often use a one-hot encoding approach, treating object classes as discrete and unrelated. This method maximizes inter-class distances but ignores the semantic relationships between different object types. The authors therein introduce a new formulation where they replace these learnable class prototypes with fixed object type embeddings derived from knowledge graphs. Specifically, the object detector learns to map visual features into a semantic embedding space, using either word embeddings (like GloVe)~\cite{PenningtonSM14} or embeddings derived directly from knowledge graphs using any standard KGE models.

Given the visual features of an object in the image, \( \mathbf{b}_i \in \mathbb{R}^D \), and the class prototypes \( \mathbf{t}_c \in \mathbb{R}^D \) for each object category \( c \), they propose nearest-neighbor classification based on the similarity between these visual embeddings and class prototypes. They explore different distance metrics, including the Manhattan distance, cosine similarity.
They then interpret the negated distance as a similarity measure \( \text{sim}(\mathbf{b}_i, \mathbf{t}_c) = 1 - d_{\text{cos}}(\mathbf{b}_i, \mathbf{t}_c)/2 \) and define the embedding loss as,

\[
L_{\text{embed}}(\mathbf{b}_i, c_i) = -\log \left( \frac{\exp(\text{sim}(\mathbf{b}_i, \mathbf{t}_{c_i})/\tau)}{\sum_{c=1}^{C} \exp(\text{sim}(\mathbf{b}_i, \mathbf{t}_c)/\tau)} \right),
\]

where \( c_i \) is the ground truth class of the object, and \( \tau \) is a temperature parameter that scales the similarity vector.
In experiments, this approach demonstrated more semantically grounded misclassifications, meaning the errors made by the model were often more contextually appropriate. Additionally, their evaluation on benchmark datasets showed that KGE-based models matched or even outperformed traditional one-hot methods, particularly in challenging object detection benchmarks. The use of KGE models also reduced the likelihood of inter-category confusions, leading to improved robustness in object detection tasks.

Yang et al.~\cite{YangHXL22} uses KG augmentation to suppress noise and generate more robust item representations for recommender system. Essentially they propose a novel Knowledge Graph Contrastive Learning (KGCL) framework designed to enhance the robustness of item representations in recommender systems by leveraging contrastive learning on KGs. The KGCL framework aims to address two challenges: the long-tail distribution of entities within KGs and the presence of noisy, topic-irrelevant connections that can affect the quality of information aggregation.
To suppress noise in the KG, KGCL first performs an augmentation process. This involves generating two stochastic views of the KG structure, denoted as $\mathcal{G}_1$ and $\mathcal{G}_2$, by randomly dropping a portion of edges in the original KG $\mathcal{G}$. Let $\mathbf{e}_i$ represent the embedding of entity $i$ in the KG. In each view, the KG aggregation function $f_\theta$ integrates the neighborhood information to generate new entity embeddings:
\[
\mathbf{e}_i^{(v)} = f_\theta(\{\mathbf{e}_j : j \in \mathcal{N}_i^{(v)}\}),
\]
where $\mathcal{N}_i^{(v)}$ denotes the set of neighbors of entity $i$ in view $v \in \{1, 2\}$. This augmentation is designed to introduce perturbations in the neighborhood structure, enabling the model to learn robust item embeddings by contrasting different views.

The core of the KGCL framework is contrastive learning, which encourages the model to maximize agreement between the embeddings of the same entity across the two views while minimizing the similarity with other entities. For each entity $i$, KGCL defines the contrastive objective using the normalized temperature-scaled cross-entropy (NT-Xent) loss,
\[
\mathcal{L}_i = -\log \frac{\exp(\text{sim}(\mathbf{e}_i^{(1)}, \mathbf{e}_i^{(2)}) / \tau)}{\sum_{j \in \mathcal{I}} \exp(\text{sim}(\mathbf{e}_i^{(1)}, \mathbf{e}_j^{(2)}) / \tau)},
\]
where $\text{sim}(\mathbf{e}_i^{(1)}, \mathbf{e}_i^{(2)})$ denotes the cosine similarity between the embeddings $\mathbf{e}_i^{(1)}$ and $\mathbf{e}_i^{(2)}$, $\tau$ is a temperature parameter, and $\mathcal{I}$ is the set of all entities. This loss function encourages the model to bring the embeddings of the same entity in different views closer while pushing apart the embeddings of different entities, thereby making the representations more robust to noise.
Once the robust item embeddings are learned, they are used to enhance the recommendation model. Given a user $u$ and an item $i$, the prediction score for user-item interaction is calculated as, $\hat{y}_{ui} = g(\mathbf{e}_u, \mathbf{e}_i)$,
where $\mathbf{e}_u$ and $\mathbf{e}_i$ are the embeddings of user $u$ and item $i$, respectively, and $g$ is a scoring function, such as a dot product. The final objective function of KGCL combines the contrastive loss $\mathcal{L}_{CL}$ and the recommendation loss $\mathcal{L}_{rec}$ as,
$\mathcal{L} = \mathcal{L}_{rec} + \lambda \mathcal{L}_{CL}$,
where $\lambda$ is a hyperparameter controlling the balance between the recommendation task and the contrastive learning objective. Through this contrastive learning mechanism, KGCL suppresses noise in the KG and enhances item representations, resulting in improved robustness for the recommendation model, especially in scenarios with sparse user-item interactions.

Radtke et al.~\cite{radtke2023increasing}  propose using KGs to enhance deep learning models for fault diagnostics in prognostics and health management (PHM). They introduce a KG-enhanced deep learning approach to incorporate domain-invariant knowledge, improving model robustness and generalization.
The method leverages the structure of KGs to encode semantic information hierarchically and combines this with supervised contrastive learning to create a more stable feature representation. Experimental results demonstrate that this approach increases the model's ability to handle domain shifts, making fault diagnostics more resilient across varying conditions.

\subsubsection{Robustness of KG-based Systems}

There are furthermore some works that do not directly discuss the robustness of the KGE models, however, consider the KG-driven systems, such as knowledge-grounded dialogue system~\cite{WangQWLHL024}, entity linkning~\cite{MaoWWL21}, cross-lingual entity alignment~\cite{pei2020rea}.
Wang et al.~\cite{WangQWLHL024} introduce an entity-based contrastive learning framework, named EnCo, to enhance the robustness of Knowledge-Grounded Dialogue (KGD) systems. Given a dialogue context $C = \{u_1, u_2, \dots, u_{n-1}\}$ consisting of utterances $u_i$ and an external knowledge set $K = \{(h_1, r_1, t_1), \dots, (h_m, r_m, t_m)\}$ comprising knowledge triples where $h_i$, $r_i$, and $t_i$ represent the head entity, relation, and tail entity respectively, the goal of a KGD system is to generate a response $u_n$ based on $C$ and $K$. The authors aim to enhance the robustness of KGD models to handle real-world perturbations, including semantic-irrelevant (e.g., misspellings, paraphrasing) and semantic-relevant (e.g., incorrect entity replacements) perturbations. To this end, they leverage contrastive learning to improve robustness by constructing positive and negative samples and training the model to recognize semantic similarities and differences.

Mao et al.~\cite{MaoWWL21} propose a robust entity alignment (EA) method that by tackling 3 aspects, namely, inefficient graph encoders, the need for negative sampling, and catastrophic forgetting in semi-supervised learning. To improve the graph encoders therein they use relational attention to update entity features as follows,
    \[
    h_i^{(l+1)} = \sigma \left( \sum_{j \in N_i} \alpha_{ij} \cdot \phi(h_j, r) \right),
    \]
where $N_i$ are neighbors of entity $i$, $\alpha_{ij}$ is the attention coefficient, and $\phi$ is a relation-specific function.
Furthermore, the authors prove that negative samples are unnecessary in EA. It adopts a symmetric negative-free alignment loss to align entity pairs without generating negative samples thereby removing the need for negative samples, aligning entity pairs with the loss,
    \[
    L = -\sum_{(i, j) \in S} \left[ \text{sim}(h_i, h_j) + \text{sim}(h_j, h_i) \right],
    \]
where $\text{sim}$ measures similarity between current and initial embeddings. Finally, to mitigate catastrophic forgetting, the approach stores previously learned embeddings and selectively reviews them during each training iteration. This approach allows the model to maintain alignment accuracy without retraining on all previous data.The evaluation has shown state-of-the-art results with improved robustness, making it efficient for large-scale knowledge graphs.

Li et al.~\cite{LiSZZ20} proposes an enhanced knowledge graph embedding method to improve the robustness of translation-based knowledge graph embedding models by introducing semantic constraints on the relations between entities. KRC introduces relational constraints on both the subject ($h$) and object ($t$) entities. These constraints are modeled using a standardized Euclidean distance:
    \[
    C_s(h^*) = \left\| \frac{h^* - \mu_s(r)}{\sigma_s(r)} \right\|_2, \quad C_o(t^*) = \left\| \frac{t^* - \mu_o(r)}{\sigma_o(r)} \right\|_2,
    \]
    where $\mu_s(r)$ and $\mu_o(r)$ are the relation-specific means, and $\sigma_s(r)$ and $\sigma_o(r)$ are standard deviations. The relational constraints are thereafter incorporated into the traditional score function of the translation-based KGE model as:
    \[
    \phi^{\text{KRC}}_{\Theta}(h, r, t) = \|h^* + r - t^*\|_p + \lambda_1 C_s(h^*) + \lambda_2 C_o(t^*),
    \]
    where $\lambda_1$ and $\lambda_2$ are hyperparameters to balance the weight of the relational constraints. Finally, to train the model, the paper introduces a novel soft margin-based ranking loss that considers the semantic distance between positive and negative triplets:
    \[
    L = \sum_{\xi \in \Delta} \sum_{\xi' \in \Delta'} \left[ \phi^{\text{KRC}}_{\Theta}(\xi) - \phi^{\text{KRC}}_{\Theta}(\xi') + \gamma (1 + \nu(\xi, \xi')) \right]_+,
    \]
    where $\Delta$ and $\Delta'$ are the sets of positive and negative triplets, $\gamma$ is the margin, and $\nu(\xi, \xi')$ is a weight function that reflects the semantic distance between triplets.
The authors claim that it results in more accurate knowledge graph completion and improved predictive performance, as validated through extensive experiments.

Pie et al.~\cite{pei2020rea} propose an approach to make robust cross-lingual entity alignment between KGs by incorporating noise detection into the alignment process using a generative adversarial network (GAN)-based approach~\cite{GoodfellowPMXWOCB14}. The model consists of a Graph Neural Network (GNN) for entity embedding and a Generative Adversarial Network (GAN) for noise detection. The GAN therein consists of a generator $G$ and a discriminator $D$. The generator generates fake entity pairs, while the discriminator assigns a trust score $T(e_1, e_2)$ to distinguish correct and noisy pairs. 
To align entities across KGs, a margin-based ranking loss is used to bring correct entity pairs closer together and push noisy pairs further apart.

The total training objective combines the discriminator loss, generator loss, and weighted alignment loss:
\[
\mathcal{L}_{total} = \mathcal{L}_D + \lambda \cdot \mathcal{L}_G + \alpha \cdot \mathcal{L}_{weighted}
\]
where $\lambda$ and $\alpha$ control the balance between noise detection and entity alignment. REA integrates GNN-based entity embeddings with GAN-based noise detection to improve robustness in cross-lingual entity alignment. The adversarial setup helps detect noisy labels and improve the accuracy of alignment by weighting the alignment loss with a trust score derived from the discriminator.

\section{Challenges and Future works}\label{sec:challenges}

Future research in the domain of resilience on knowledge graphs and KGE models presents a number of possibilities to improve different aspects of resilience that we defined in this work. We can envisage works aiming at developing KGE models considering generalization consistency, distribution adaptation, and performance consistency, amongst others. We describe future work directions in more detail in the following.

\begin{description}
\item[Distribution shift and consistence performance.] One promising avenue for future work is the development of resilient KGE models that can adaptively adjust to changes in the underlying data or graph structure. Traditional KGE models often assume static or stationary environments, which may not hold in dynamic or evolving KGs. Future research could explore dynamic embedding techniques that continuously update entity and relation embeddings to capture temporal or contextual changes, for instance, when new entities or relations are being added to the KG. Additionally, investigating the integration of uncertainty modeling and probabilistic reasoning mechanisms into KGE models could enhance their resilience to noisy or uncertain data. There are some works quantifying the uncertainty of KGE models such as~\cite{ChenCSSZ19,LiuZZW24}.  The works in~\cite{ChenCSSZ19} came up with the {\em probabilistic soft logic} to generate confidence scores capturing the structural and assertional uncertainties. When new entities and relations are added to the KG (i.e., when a shift of distribution occurs), the underlying KGE models can give predictions with some confidence measure to denote the uncertainty in predictions. In~\cite{LiuZZW24}, the authors define a KG as uncertain when each assertion is associated with a confidence score, which is also given to KGE algorithms during the embedding computation process. These works have laid the foundation for making KGE models aware of distribution shifts by considering uncertainty as part of the prediction of KGE models. However, many such works are still needed. Additionally, depending on the domain, the allowable shift (as defined in Equation~\ref{eq:hd}, $\mathcal{H}(\mathcal{D}_{\mathcal{G}}, \mathcal{D}_{\mathcal{G'}})$) should also be precisely defined. To this end, KGE models should consider different possible distribution shifts, such as node, edge, and graph distribution shifts.

\item[Adversarial and non-adversarial robustness.] Another possible research direction is that of resilience of KGE models against adversarial attacks and manipulations, i.e., developing KGE models that are adversarially robust. As mentioned beforehand, real-world KGs might suffer from adversarial attacks where adversaries may attempt to exploit vulnerabilities in the KG or KGE models to inject false information, manipulate inference results, or disrupt system functionality. There have already been some works to this end, however, all of them focus on developing methods to perform targeted attacks, i.e., considering a specific fact to add or remove from the KG and thereby making the KGE model learn based on the attackers' goal. To this end, only the KG has been considered as a possible attack surface. However, there can be other possibilities, for instance, the parameters of the already trained KGE model can be attacked. Such kind of attacks are often prevalent in the ML domain and termed Trojan attacks~\cite{LiuMALZW018,LiuM0020,abs-2202-07183} where the attacker aim to make the model learn their objective either by generating inputs with certain {\em triggers}, or by changing the already trained model's parameters. For KGE models, such Trojan attacks could correspond to the modification of the entries of learned embedding vectors so as to achieve a specific attacker's objective. 
Apart from considering targeted attacks by taking into account different attack surfaces, it would also be needed to consider performing non-targeted attacks~\cite{abs-2112-00639}, where the idea is to simply disrupt the performance of the underlying KGE models by introducing noise in the KGs or in the KGE models.  

Additionally, the targeted attacks so far have been considered only for a specific type of task, namely link prediction tasks. KGE models are used in many critical downstream application tasks~\cite{DaltonDA14,FerrucciBCFGKLMNPSW10,WangZZLXG19}, and hence, more research is needed to understand how to perform adversarial attacks on such KGE-based tasks. This basically opens up a number of different attack surfaces along with the need to explore different attack dimensions, including non-targeted attacks.

While several works considered adversarial attacks on KGE models, a much-needed direction to be focused on is the development of defence mechanisms against such attacks that can detect and mitigate adversarial attacks in real-time, thereby enhancing the overall robustness of KG-based applications. This would include developing graph-based anomaly detection algorithms to identify and mitigate adversarial attacks or abnormal patterns in the KG, performing adversarial training of KGE algorithms, developing certified guaranteed methods to build robust KGE models, and so on. Furthermore, defence mechanisms should be extended to combat non-targeted attacks, effectively addressing noise, inconsistencies, or incompleteness inherent in KGs. This entails the creation of robust data integration and ensemble algorithms capable of handling diverse and noisy information from various sources. Moreover, exploring techniques for automated error detection, correction, and data validation within KGs could significantly enhance their quality and reliability over time. 

Recently, the works to combine large language models (LLMs) with KGEs are gaining popularity~\cite{PanRKSCDJO0LBMB23}. There is a potential that by augmenting LLMs to KGEs, one could achieve improved robustness. Leveraging the semantic richness of natural language representations encoded in LLMs, such as BERT~\cite{DevlinCLT19} or GPT~\cite{brown2020language}, may enhance the understanding and representation of entities and relations in the KG. This integration could potentially mitigate the impact of noisy or incomplete KGs on downstream tasks.
Conversely, incorporating KGEs into LLMs could offer benefits such as enriched context-aware embeddings for downstream natural language understanding tasks. By incorporating structured knowledge from KGs, LLMs could gain a deeper understanding of entity relationships and enable more informed decision-making.

\end{description}

Finally, resilience has already been vastly explored in fault-tolerant systems, therefore, interdisciplinary approaches that draw insights from fields such as network science, complex systems theory, and resilience engineering could provide valuable perspectives and methodologies for enhancing the resilience of KGs and KGE models. By leveraging principles from these domains, researchers can develop holistic, multi-faceted strategies for improving the reliability and robustness of KG-based systems in diverse application domains.

\section{Conclusion}
\label{sec:conclusion}
In this work, we explored the resilience of knowledge graph embedding models, addressing their ability to withstand and adapt to various challenges such as noise, adversarial attacks, and dynamic changes in the underlying knowledge graphs. While significant research has been conducted on robustness, particularly adversarial robustness, there is a pressing need to consider a more comprehensive notion of resilience. This broader understanding includes aspects such as generalization consistency, distribution adaption, and performance stability under diverse real-world conditions.
A key finding of this survey is that while adversarial robustness has received considerable attention, with various strategies to mitigate attacks on KGE models, resilience in non-adversarial contexts is equally critical. Models must not only defend against malicious interventions but also maintain their reliability in the presence of natural noise and inconsistencies prevalent in real-world KGs. The surveyed works on non-adversarial robustness primarily focus on mitigating the effects of noise by incorporating confidence-aware learning and enhanced negative sampling strategies. However, these approaches often overlook the dynamic nature of KGs, particularly temporal and evolving KGs, where distribution shifts are inevitable. Addressing such shifts through adaptive retraining mechanisms remains an open challenge.
Moreover, ensuring performance consistency across diverse application domains is essential. KGE models must be able to operate effectively even with incomplete input data, which is a common scenario in real-world applications. Achieving this consistency demands that future research goes beyond traditional robustness frameworks, integrating novel methodologies from graph neural networks, reinforcement learning, and explainable AI to enhance both the adaptability and transparency of KGE models.
In conclusion, while much progress has been made in improving the robustness of KGE models, a more holistic approach to resilience—incorporating adaptability, consistency, and robustness in the face of both adversarial and natural challenges—will be key to unlocking the full potential of these models in real-world, dynamic, and noisy environments.



\bibliography{ref}

\end{document}